\documentclass{article}
\newcommand{\iclrdir}{}
\usepackage{\iclrdir iclr2027_conference}
\usepackage{times}

\usepackage[utf8]{inputenc}
\usepackage[T1]{fontenc}
\usepackage{hyperref}
\usepackage{url}
\usepackage{booktabs}
\usepackage{amsfonts}
\usepackage{nicefrac}
\usepackage{microtype}
\usepackage{enumitem}
\usepackage{xcolor}
\usepackage{graphicx}
\graphicspath{{\iclrdir plots/}}
\usepackage{amsmath}
\usepackage{amssymb}
\usepackage{mathtools}
\usepackage{amsthm}
\usepackage{float}
\usepackage{algorithm}
\usepackage{\iclrdir algorithmic}
\hypersetup{hypertexnames=false}
\usepackage[capitalize,noabbrev]{cleveref}

\providecommand{\theHalgorithm}{\arabic{algorithm}}

\theoremstyle{plain}

\theoremstyle{definition}

\theoremstyle{remark}

\title{How Much Information Can We Fit inside a Transformer?}

\author{%
\textbf{Amirhossein Basareh}\thanks{Equal contribution.}\hspace{2em}\textbf{Rishabh Sanjay}\footnotemark[1]\hspace{2em}\textbf{Daniel de Mello}\hspace{2em}\textbf{Abulhair Saparov} \\
\textnormal{Purdue University} \\
\textnormal{\texttt{\{abasareh,sanjayr,ddemello,asaparov\}@purdue.edu}}
}

\iclrfinalcopy %
\begin{document}

\maketitle

\begin{abstract}
How efficiently can transformers store factual knowledge? The answer to this question has implications for both model design and understanding the limits of transformer-based systems. Prior studies observe a linear relationship between the number of facts memorized and the number of model parameters required, substantially worse than what theoretical analyses predict is achievable. We re-examine empirical scaling: through rigorous and systematic experiments, including extensive hyperparameter search via Bayesian optimization, we determine the smallest model that can perfectly memorize a given set of facts, and measure how this minimum size scales with the number of facts. On synthetic datasets of random, independent (subject, relation, object) triples---e.g., \texttt{(Paris, capital\_of, France)}---drawn at random with no exploitable structure, we find that memorization capacity, the size of the largest dataset that a model memorizes completely, grows exponentially with the transformer dimension. Equivalently, the minimum transformer dimension $d_{\min}$ required to memorize $N$ facts grows logarithmically in $N$. Across 11 dataset sizes spanning $N = 45$ to $49{,}920$, a lack-of-fit F-test does not reject the logarithmic fit (with an offset), but it does reject all other tested alternatives including power-law. Along the depth axis, however, at fixed width, the maximum memorizable dataset size grows sublinearly with the number of layers over the measured range. We further extend our exploration to structured data containing symmetric relations or two-hop composition. In this setting, we find that transformers exploit these regularities to further improve capacity. All code to reproduce our results is available at \url{https://github.com/Magronox/transformer_information_capacity}.
\end{abstract}

\section{Introduction}

Large language models (LLMs) exhibit a remarkable ability to memorize facts and 
associations, sometimes to the point of verbatim 
recall~\citep{carlini2021extracting, carlini2023quantifying, petroni2019language}. 
As transformer models scale to trillions of parameters, critical questions 
arise: What is the maximum number of associations a model of given size 
can memorize? How does this capacity scale with model size? 
Answering these questions enables practitioners to select appropriate architectures/hyperparameters for knowledge-intensive tasks and predict when transformers will successfully memorize versus fail to learn training data due to insufficient capacity~\citep{li2025misfitting}. Our central research questions are: (1) how does the minimum transformer model dimension required to perfectly memorize $N$ facts scale with $N$? and (2) at fixed model dimension, how does the maximum number of memorizable facts scale with the number of layers $L$?

Prior empirical studies of transformer memorization measure how many bits of factual knowledge are learned per parameter. \citet{allenzhu2024physics} train GPT-2, LLaMA, and Mistral models on synthetic biographies composed of (name, attribute, value) tuples and establish, through an information-theoretic bit-complexity analysis of their trained model, that sufficiently trained transformers store ${\sim}2$ bits of information per parameter (i.e., linear scaling). \citet{morris2026how} train GPT-style models on synthetic random sequences and report ${\sim}3.6$ bits per parameter. \citet{lu2024scaling}, training Qwen-style LLMs directly on relational triples, report approximately linear scaling between fact capacity and number of parameters, estimating that memorizing Wikidata would require ${\sim}1000$ billion parameters; however, their measurements use 50--200 training epochs while their own analysis indicates that capacity saturates only at around 1000 epochs and hyperparameters are only partially tuned at different model sizes.

Theoretical work has established bounds on transformer memorization capacity. The result most directly relevant to our work is \citet{kajitsuka2025}, who study the next-token prediction setting (a single label per input sequence, equivalent to ours) and prove that $\tilde{O}(\sqrt{N})$ parameters are sufficient for a transformer to memorize $N$ facts, and that $\Omega(\sqrt{N})$ parameters are necessary for memorizing every $\{0,1\}$-labeling of $N$ facts; these upper and lower bounds are the same up to log factors and are much better than the linear scaling observed in empirical studies. Other upper-bound constructions for related settings give larger parameter counts \citep{kajitsuka2024, madden2024, mahdavi2023memorization, kim2023provable} while Hopfield-network and Johnson--Lindenstrauss theoretical results predict transformer dimension needs to scale logarithmically with $N$ for memorization in idealized settings; we discuss both in Section~\ref{sec:related}. Our results show that the standard fixed-depth softmax transformer can memorize $N$ facts with the transformer dimension scaling as $\log N$.

We believe that the gap with prior empirical work has two main sources. Prior studies that report a linear relationship between parameter count and the number of memorized facts \citep{lu2024scaling, morris2026how} use limited hyperparameter search and limited training compute, both of which prevent the model from reaching its maximum information capacity at larger dataset sizes (Section~\ref{sec:related}). \citet{allenzhu2024physics} demonstrate that the number of times each fact is seen during training (i.e., \emph{exposure}) directly governs measured capacity, confirming that insufficient training duration causes underestimation of the model's information capacity. 

In this work we show that standard transformers empirically achieve substantially better scaling than the linear relationship between number of parameters and number of facts reported in prior empirical work \citep{allenzhu2024physics, lu2024scaling}. We train transformers on synthetic relational triples~\citep{bordes2013translating} with uniformly random structure, since we aim to test the model's memorization capability while controlling for generalization. We find that the number of facts $N$ that a transformer can memorize\footnote{I.e., the model achieves 100\% training accuracy in predicting the object given the subject and relation.} grows \emph{exponentially} with transformer dimension. Equivalently, the minimum transformer dimension $d_{\min}$ required to memorize $N$ facts grows logarithmically in $N$. In this setting, the logarithmic form is the only form that is not rejected by a lack-of-fit F-test~(Section~\ref{sec:exp-width}).
At fixed width (i.e., transformer dimension) $d = 8$ and a fixed training budget, the maximum memorizable dataset size grows with depth (i.e., number of layers), but sublinearly; a lack-of-fit F-test rejects linear, $N = aL$ ($p = 7.8\times10^{-6}$), affine, $N = aL + b$ ($p = 0.0025$), and superlinear forms ($p \leq 1.2\times10^{-5}$), while a logarithm and a power-law with exponent 0.65 are not rejected, and our data do not distinguish between them (Section~\ref{sec:exp-depth}).

In addition, we extend our analysis beyond unstructured uniformly-distributed data. For data with logical structure, such as data containing symmetric relations ($r$ is symmetric if $(s, r, o)$ then $(o, r, s)$ for any $o$, $s$) and two-hop compositional relations (if $(a, r_1, b)$ and $(b, r_2, c)$ then $(a, r_3, c)$ for any $a$, $b$, $c$), we find that transformers exploit these regularities to achieve improved scaling. To probe whether \emph{distribution skew} alone affects capacity---a relevant question because real-world fact distributions are typically Zipf-like rather than uniform---we additionally test power-law-distributed data, finding that they show similar scaling behavior to uniformly-distributed data.

Our key contributions are:
\begin{itemize}[noitemsep, topsep=4pt, leftmargin=2em]
    \item We demonstrate that memorization capacity grows exponentially with the transformer dimension: at fixed depth $L=1$, the minimum transformer dimension $d$ needed to memorize $N$ facts grows logarithmically in $N$, and a lack-of-fit F-test rejects the power-law and other alternatives (Section~\ref{sec:exp-width}, Table~\ref{tab:width-fits}).
    
    \item At fixed width and a fixed training budget, capacity grows sublinearly with the number of layers: linear and superlinear growth are rejected, and the sublinear forms we tested are not distinguished (Section~\ref{sec:exp-depth}, Table~\ref{tab:depth-forms}).
    
    \item We show that the gap with prior empirical work stems from 
    training methodology: both insufficient training duration and limited 
    hyperparameter optimization lead to suboptimal scaling trends (Section~\ref{sec:ablation}).
    
    \item We characterize how relational structure in the data and the shape of the data distribution affect memorization. Structured data (symmetric relations and two-hop compositions, defined in Section~\ref{sec:datasets}) leads to improved scaling by exploiting regularities, and power-law-distributed data shows a similar scaling behavior to uniformly distributed data (Section~\ref{sec:exp-width}, Figure~\ref{fig:min-dim-consolidated}).
\end{itemize}

These findings suggest that LLMs may have substantially 
greater knowledge capacity than previous empirical estimates indicate, 
with implications for understanding how transformers store 
knowledge~\citep{geva2021transformer} and for predicting capacity 
requirements at scale.

\section{Related Work}
\label{sec:related}

\paragraph{Scaling Laws.}
Empirical scaling laws characterize how language model performance varies with computational resources. \citet{kaplan2020scaling} established power-law relationships between loss and model parameters, dataset size, and training compute; \citet{hoffmann2022chinchilla} revised their compute-allocation recommendations, showing parameters and training tokens should scale equally. \citet{henighan2020scaling} extended scaling laws to multimodal domains. Extensive work on scaling laws has since followed, and we refer the reader to a survey \citep{li2025misfitting}. These scaling laws characterize average performance (cross-entropy loss) rather than memorization capacity.

\paragraph{Empirical Memorization Studies.}Earlier work by \citet{tirumala2022memorization} showed that larger transformer models memorize training data faster and can memorize a larger fraction of the training data before overfitting, i.e., before their validation loss begins to increase. Three recent studies measure transformer memorization capacity, with complementary methodologies and findings. \citet{lu2024scaling} train Qwen-style LLMs on relational triples and report approximately linear scaling between parameters and fact capacity, estimating that memorizing Wikidata would require ${\sim}1000$ billion parameters; their scaling measurements use 50--200 training epochs, while their own analysis indicates that the fraction of facts recalled correctly continues to improve up to 1000 epochs, suggesting that longer training might yield better scaling behavior. \citet{allenzhu2024physics} introduce an information-theoretic framework, training GPT-2, LLaMA, and Mistral models on synthetic biography datasets composed of (name, attribute, value) tuples---closely analogous to the relational triples studied by \citet{lu2024scaling} and in our work. They establish ${\sim}2$ bits/param for sufficiently trained transformers. They further show that the number of times each fact is seen during training critically affects measured capacity: with 1000 exposures per fact, transformers achieve 2 bits/param, but with only 100 exposures, they achieve 1 bit/param. \citet{morris2026how} estimate capacity for GPT-style models at ${\sim}3.6$ bits per parameter. Crucially, the bits/param measurements of \citet{allenzhu2024physics} and \citet{morris2026how} are measurements for sufficiently trained transformers for a specific dataset size, not assertions about \emph{how capacity scales with dataset size} more broadly. Our work is complementary: rather than reporting a bits-per-parameter estimate for a single dataset size, we fit a precise functional form $d_{\min}(N)$, the minimum transformer dimension that achieves perfect memorization, across a wide range of values of $N$, finding that $d_{\min}$ grows logarithmically in $N$ across this range.

\paragraph{Theoretical Bounds on Memorization Capacity.}
Theoretical analyses of transformer memorization build on contextual mapping constructions~\citep{DBLP:conf/iclr/YunBRRK20}. These results are typically worst-case bounds over arbitrary labelings, and assumptions vary across settings (e.g., the prediction task, label space, and separation conditions on inputs). Most directly relevant to our setting, \citet{kajitsuka2025} prove that $\tilde{O}(\sqrt{N})$ parameters are sufficient to memorize $N$ sequences with a single label each in the next-token prediction setting, and that this bound is tight up to log factors. Earlier upper-bound constructions for related but distinct settings give different parameter-count scalings: (1) $O(d\ell N + d^2)$ parameters for sequence-to-sequence prediction with one-layer transformers with transformer dimension $d$ and input sequence length $\ell$~\citep{kajitsuka2024}, (2) $\smash{\tilde{O}}(d + \ell + \smash{\sqrt{\ell N}})$ parameters for sequence-to-sequence prediction with $O(\ell)$ layers~\citep{kim2023provable}, (3) $\Theta(\zeta  K)$ parameters for next-token \emph{distribution} prediction over $K$ distinct contexts with a vocabulary of size $\zeta$~\citep{madden2024}, and (4) $\Theta(Hd^2)$ parameters for a multi-head attention layer with $H$ heads to memorize $\Omega(H\ell)$ linearly independent inputs~\citep{mahdavi2023memorization}.

\citet{nichani2024understanding} study a synthetic factual-recall task with \emph{fixed random embeddings} and prove that a single-layer transformer reaches perfect accuracy when either its self-attention or its MLP parameters scale linearly with the number of facts $N$ (up to log factors). Our embeddings are jointly trained, and the number of non-embedding parameters grow as $(\log N)^2$ in $N$, rather than the linear $P \sim N$.

\citet{nichani2024understanding} further prove that a single self-attention layer followed by an MLP stores facts at a rate proportional to its parameter count, and \citet{chen2024what} find on synthetic sequence tasks that one attention layer suffices for memorization while reasoning and generalization need two or three layers. Neither result predicts what gradient training can achieve when layers are added at a fixed width and a fixed training budget. We measure this in our depth experiment.

Two theoretical results predict that storing $N$ items in a $d$-dimensional space is achievable when $d \sim \log N$. The Johnson--Lindenstrauss lemma \citep{johnson1984extensions} establishes that $\Theta(e^{cd})$ near-orthogonal unit vectors fit in $d$-dimensional Euclidean space. Equivalently, $d = O(\log N)$ suffices to fit $N$ such vectors. Modern Hopfield-network analyses \citep{ramsauer2020hopfield, lucibello2024exponential} show that softmax attention, viewed as an associative memory over stored patterns, has retrieval capacity exponential in the pattern dimension. Both results concern idealized settings---vector geometry and a single softmax retrieval step over fixed patterns, respectively---rather than trained transformers with FFN, residuals, and learned embeddings.%

Our results on memorization capacity scaling with respect to transformer dimension are consistent with concurrent work by \citet{ravfogel2026geometric}, who fit a logarithmic curve to the smallest embedding dimension that reaches 95\% training accuracy on random bijections. However, they do not report a statistical test and do not compare the logarithmic fit with other forms such as a power-law. Our work, on the other hand, systematically studies the scaling of memorization capacity across 11 dataset sizes and compares logarithmic, power-law and linear forms with a lack-of-fit F-test (Section~\ref{sec:exp-width}).
\section{Background and Methods}
\label{sec:methods}

Throughout, we use $N$ to denote the number of facts in the dataset, $d$ for the transformer dimension (which equals the embedding dimension in our setting, by standard convention), $L$ for the number of layers, and $P$ for the transformer (non-embedding) parameter count. The total number of parameters is $P + |V| \cdot d$, where $|V| \cdot d$ is the size of the embedding table; in our experimental setup $|V|$ scales with $N$.

Our goal is to investigate how transformer memorization capacity scales with model dimension (width) and number of layers (depth) under realistic training constraints.

We measure capacity along two axes: (1) given a dataset, we find the smallest dimension $d^*$ that achieves perfect memorization (100\% accuracy); (2) given a model, we find the largest dataset size $N^*$ it can perfectly memorize. We optimize training hyperparameters in each setting to maximize measured capacity. For detailed mathematical formulation, see Appendix~\ref{app:optimization}.
 
We make a simplifying assumption that all facts have the form $(s, r, o)$, where $s$ is the subject entity, $r$ is the relation, and $o$ is the object entity. For example, the fact ``Paris is the capital of France'' can be expressed as $(\texttt{Paris}, \texttt{capital\_of}, \texttt{France})$. This $(s, r, o)$ representation is widely adopted in empirical and theoretical studies of memorization and factual recall in language models \citep{bordes2013translating, petroni2019language, dai2022knowledge, meng2022locating, hernandez2024lineardecodingm, allenzhu2024physics, lu2024scaling, nichani2024understanding}. In this controlled setting, on unstructured data, the model can only achieve high accuracy by memorizing facts. The $(s, r, o)$ abstraction corresponds to these representations. Further, it is easy to generate an arbitrary number of $(s,r,o)$ triples with a high degree of control over their distribution and structure. Studying the ability of transformers to memorize triples facilitates our measurement of the fundamental capacity of the associative memory mechanisms that transformers employ for factual knowledge, enabling us to probe the model's memorization capacity independently of its information extraction ability.
 
\subsection{Task Formulation}
 
We measure memorization capacity using knowledge graph triples $(s, r, o)$ where the subject $s$ and relation $r$ together uniquely specify the object $o$. Each fact query is tokenized as a sequence of two input tokens $(s, r)$ and the model must learn to map each input to its unique corresponding output token $o$. This task requires perfect memorization of all $N$ associations in the dataset.

\paragraph{Definition (memorization and capacity).} A trained model \emph{memorizes} a fact $(s, r, o)$ if, given the two-token input $(s, r)$, its highest-probability output token is $o$ (i.e., greedy decoding). A model \emph{memorizes a dataset} of $N$ facts if it memorizes every fact---that is, it reaches 100\% training accuracy. \emph{Capacity} is defined as the largest dataset size that a given model can memorize completely. The transformer width and depth scaling experiments measure this same quantity in two different ways. Let $\mathrm{Train}(d, L, N; \theta)$ denote the model obtained by training a transformer of dimension $d$ and depth $L$ on a dataset of $N$ facts with hyperparameters $\theta$ (Table~\ref{tab:hparam-search}), until it reaches 100\% training accuracy or exhausts the training budget. We consider two methods for measuring the capacity of the transformer: (1) %
We fix the dataset size $N$ and search for the smallest transformer model dimension $d$ such that $\mathrm{Train}(d, L, N; \theta)$ memorizes the dataset for some setting of hyperparameters $\theta$ (Eq.~\ref{eq:width-obj}); this finds the smallest model whose capacity is at least $N$. %
(2) We fix the model and search for the largest $N$ such that $\mathrm{Train}(d, L, N; \theta)$ memorizes the dataset (Eq.~\ref{eq:depth-obj}); this measures the model's capacity directly. For brevity, we focus on the first method for the width scaling experiments, where we fix $L=1$, but we demonstrate that both methods yield the same results in Appendix~\ref{sec:exp-maxn-width}. For the depth scaling experiments, since $d$ is fixed, the first method is not feasible and so we apply the second method. All reported values use the hyperparameter configuration selected by the search described in Appendix~\ref{app:hyperopt}, with 60 completed trials per learning-rate schedule. %

\subsection{Model Architecture}
 
We use the GPT-2~\citep{radford2019language} architecture, with the following differences: a single attention head ($H = 1$), the Swish activation function $x \cdot \mathrm{sigmoid}(x)$ in place of GELU, no dropout, and no positional embeddings;
since every input has the fixed form $(s, r)$, position is
implicit in the input structure. In preliminary experiments, models
trained with and without learned positional embeddings required the same
minimum dimension (Appendix~\ref{sec:Appendix-results}). The vocabulary has $|V| = n + m$ tokens, one for each entity and relation, in a single embedding table of size $|V| \times d$, shared between input and output.

Following standard convention~\citep{kaplan2020scaling}, the embedding dimension is $d$, so parameters are split into embedding ($|V| \cdot d$) and non-embedding ($\Theta(L d^2)$, attention and feed-forward weights); the latter is $P = L(12d^2 + 13d) + 2d$ for our single-head architecture, which equals $12d^2 + 15d$ at $L = 1$. We report $P$ and the embedding parameters separately throughout, since the embedding table is determined by vocabulary and $P$ is determined by the architecture alone.

\subsection{Hyperparameter Optimization}
 
We optimize training hyperparameters using Bayesian optimization (BO) with tree-structured Parzen estimator (TPE) sampling via the \texttt{Optuna} library~\citep{NIPS2011_86e8f7ab, akiba2019optuna}. We stratify the search by learning-rate schedule (\texttt{linear} or \texttt{cosine}): a separate Bayesian optimization is run for each schedule and the best across schedules is reported. The search space is given in Table~\ref{tab:hparam-search} (Appendix~\ref{app:hyperopt}); each candidate is evaluated on three seeds, with 60 completed trials per schedule for every study.

\subsection{Training Protocol}
 
We train in \texttt{bfloat16} mixed precision using the Adam optimizer~\citep{kingma2015adam} to minimize cross-entropy loss with the $L_2$ weight decay parameter obtained via Bayesian optimization. Training is full-batch: every fact appears in every iteration. We cap training at 40{,}000 iterations (excluding warmup)---up to 40{,}000 exposures per fact---based on our convergence analysis (Appendix~\ref{sec:Appendix-results}), with early stopping only after 100\% accuracy holds for 50 consecutive iterations, i.e., after memorization is complete. That is 40$\times$ the 1{,}000-exposure saturation regime of \citet{allenzhu2024physics}.

\subsection{Capacity Search Procedures}
\label{sec:dim-search}
A training run \emph{fails} if it does not reach 100\% training accuracy and maintain it for 50 consecutive iterations within the 40{,}000-iteration budget, and \emph{succeeds} otherwise. For the experiments where we aim to find the smallest model dimension that leads to successful memorization, we perform a search to find $d^*$---the smallest dimension at which training succeeds---for each hyperparameter configuration. %
To this end, we perform a two-phase search: In the first phase, we repeatedly perform geometric halving, starting from $d=32$, %
until the first failure (or doubling $d$ until the first success if the starting estimate fails). In the second phase, we perform binary search to pinpoint $d^*$. For the experiments where we search for the largest dataset that leads to successful memorization, we similarly perform the same two-phase search procedure to find $N^*$---the largest dataset size at which training succeeds. The full procedure is in Appendix~\ref{app:hyperopt}.

\subsection{Dataset Construction}
\label{sec:datasets}
 
For all experiments, we use synthetic datasets comprising $(s, r, o)$ triples. We generate both unstructured and structured datasets to investigate whether transformers can exploit structure to achieve higher memorization capacity. To create a dataset, we specify two parameters: the number of entities $n$ and the number of relations $m$.
 
We generate four datasets: \texttt{Uniform}, \texttt{Symmetric}, \texttt{2Hop}, and \texttt{PowerLaw}.

To remain consistent with our deterministic full-batch methodology, each batch contains every fact. For \texttt{Uniform}, \texttt{Symmetric}, and \texttt{2Hop}, each fact appears once per batch. For \texttt{PowerLaw} (Zipf with exponent $\alpha = 1.0$), each fact $i$ appears $\max(1, \lceil k\, p_i \rceil)$ times per batch, where $p_i$ is its Zipf probability and the multiplier $k = 1/p_N$ is chosen so that the rarest fact appears exactly once per batch. Frequent facts therefore appear more times per batch than rare ones; full details for generating \texttt{PowerLaw} are in Appendix~\ref{app:powerlaw}. We conduct experiments with $n \in \{15, 30, 60, 120, 260, 520, 1040, 2080, 4160, 8320, 16640\}$ entities and $m=3$ relations across all dataset families (\texttt{Symmetric} up to $n = 4160$ and \texttt{2Hop} up to $n = 8320$). For \texttt{PowerLaw}, we only consider $n \in \{15, 30, 60, 120, 260, 520, 1040, 2080, 4160\}$ since $k$ grows with $n$ and the batch becomes prohibitively large otherwise.

\textbf{Uniform Dataset} (unstructured): For \texttt{Uniform}, we create a dataset such that, for any subject $s \in \{1, 2, \cdots, n\}$ and for any relation $r \in \{1, 2, \cdots, m\}$, there is a triple in the dataset $(s,r,o)$ where the object $o$ is uniformly sampled from the set $\{1, 2, \cdots, n\}$. For any subject-relation pair $(s,r)$, we have a unique object $o$. Therefore, the dataset size is $m n$.
 
\textbf{Symmetric Dataset} (structured): For \texttt{Symmetric}, for each relation $r$ we shuffle the $n$ entities uniformly at random and pair them in order, the first with the second, the third with the fourth, and so on (when $n$ is odd, the last entity is paired with itself). Each entity is in exactly one pair, and for each
pair $\{s, o\}$ we add both $(s, r, o)$ and $(o, r, s)$ to the dataset (a single fact if $o = s$). Every relation is therefore symmetric: if a triple $(s, r, o)$ is in the dataset, then its counterpart $(o, r, s)$ is also in the dataset (e.g., if $a$ is a sibling of $b$, then $b$ is a sibling of $a$).
 
\textbf{2Hop Dataset} (structured): For \texttt{2Hop}, we use $m = 3$ relations and dedicate the last relation $r_3$ to the two-hop compositions. For relations $r_1$ and $r_2$, objects are sampled uniformly as in \texttt{Uniform}; then, for each subject $a$, we add the triple $(a, r_3, c)$, where $b$ and $c$ are the entities such that $(a, r_1, b)$ and $(b, r_2, c)$ are in the dataset. This is analogous to real-world relations that have the same multi-hop compositional property (e.g., if $a$ is the mother of $b$ and $b$ is the parent of $c$, then $a$ is the grandmother of $c$). This format preserves the uniqueness property, i.e., for any pair $(s, r)$, $o$ is unique. Step-by-step generation procedures for all dataset families are given in Appendix~\ref{sec:Appendix-algorithms} (Algorithms~\ref{alg:gen-dataset} and~\ref{alg:powerlaw-batch}).
 
\textbf{Power-law Dataset}: To probe whether non-uniform fact distributions affect memorization capacity, we sample triples from a Zipf distribution: each triple is assigned a rank $i \in\{1, 2, \ldots, nm\}$ and sampled with probability $p_i \propto i^{-\alpha}$.%

\section{Results}
 
We characterize how transformer memorization capacity scales with model dimension $d$ (width) and number of layers $L$ (depth).
 
\subsection{Width scaling: Capacity grows exponentially with transformer width}
\label{sec:exp-width}
 
\begin{figure}
    \centering
  \begin{minipage}[c]{0.47\linewidth}
    \includegraphics[width=\linewidth]{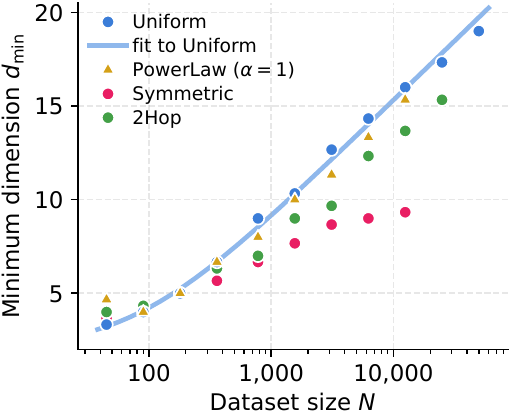}
  \end{minipage}\hfill
  \begin{minipage}[c]{0.49\linewidth}
  \caption{Minimum transformer dimension $d_{\min}$ required for complete
    memorization vs.\ dataset size $N$ (log scale) at $L=1$, mean over three
    seeds. \textbf{Uniform} (blue circles), with the logarithmic-with-offset
    fit of Table~\ref{tab:width-fits} (light blue line).
    \textbf{PowerLaw} (amber triangles; Zipf $\alpha=1.0$), tested up to
    $N=12{,}480$. \textbf{Symmetric} (pink circles), up to $N=12{,}480$, and
    \textbf{2Hop} (green circles), up to $N=24{,}960$.}
  \label{fig:min-dim-consolidated}
  \end{minipage}
\end{figure}
 
Figure~\ref{fig:min-dim-consolidated} plots the minimum transformer dimension required for complete memorization across all four dataset families at $L=1$. Two findings are visible.
 
\textbf{Exponential capacity scaling on \texttt{Uniform}.} As a general observation, across the measured range $N \in [45, 49{,}920]$ the number of non-embedding parameters per memorized fact strictly decreases as the dataset grows; the logarithmic fit below is our best quantitative description of this trend within this range. Table~\ref{tab:width-fits} compares candidate fits on the 11 \texttt{Uniform} measurements, spanning $N = 45$ to $49{,}920$ (all fits computed in log space; we compare forms with the lack-of-fit F-test for repeated measurements, \citealt{lof}, using the three seeds at each size, Appendix~\ref{app:loft}). The logarithmic-with-offset form $d = 2.756\,\log(N + 80.6) - 10.09$ achieves $R^2 = 0.997$, and is not rejected by a lack-of-fit F-test ($p = 0.67$); the power-law is rejected with or without an offset. The largest dataset sizes are what separate the two forms. Considering only dataset sizes up to $N = 6{,}240$, neither power-law is rejected ($p = 0.23$ without an offset, $0.16$ with one); with the sizes up to $N = 12{,}480$, both are rejected ($p = 0.028$ and $0.018$).

\begin{table}[!ht]
\centering
\caption{Width-scaling fits on \texttt{Uniform} ($L=1$, 11 measured points, $N = 45$ to $49{,}920$). All fits are computed in log space, with each fit (except the pure logarithmic fit) constrained so that the fitted $d$ is positive for every $N > 0$; the last column is the lack-of-fit F-test $p$-value ($p < 0.05$ rejects the functional form).}
\vspace{4pt}
\label{tab:width-fits}
\small
\begin{tabular}{llrr}
\toprule
\textbf{Functional form} & \textbf{Best fit} & $R^2$ & \textbf{Lack-of-fit $p$-value} \\
\midrule
Logarithmic\,+\,offset & $d = 2.756\,\log(N + 80.6) - 10.09$ & $0.997$ & $0.67$ \\
Logarithmic            & $d = 2.17\,\log N - 5.39$           & $0.979$ & $9.1 \times 10^{-4}$ \\
Power-law\,+\,offset   & $d = 1.38\,N^{0.259} + 0.00$       & $0.967$ & $1.7 \times 10^{-5}$ \\
Power-law              & $d = 1.38\,N^{0.259}$              & $0.967$ & $3.0 \times 10^{-5}$ \\
Linear                 & $d = 4.83 \times 10^{-4}\, N + 6.30$ & $0.565$ & $4.4 \times 10^{-16}$ \\
\bottomrule
\end{tabular}
\end{table}

\textbf{Structured datasets require smaller dimension.} For \texttt{Symmetric} (where the relation is its own inverse) and \texttt{2Hop} (where we have two-hop compositional relations), $d_{\min}$ grows more slowly than for \texttt{Uniform}. Under our full hyperparameter optimization protocol, this gap is large and consistent from $N = 360$ onward (at the three smallest dataset sizes, all families require similar dimensions). When the hyperparameter search is restricted to finding only the learning rate and batch size, as in prior empirical work, the three families require the same dimension, as we discuss in the ablation studies (Section~\ref{sec:ablation}).
 
\subsection{Depth scaling: Capacity grows sublinearly with the number of layers}
\label{sec:exp-depth}

\begin{figure}[!ht]
  \centering
  \begin{minipage}[c]{0.47\linewidth}
  \includegraphics[width=\linewidth]{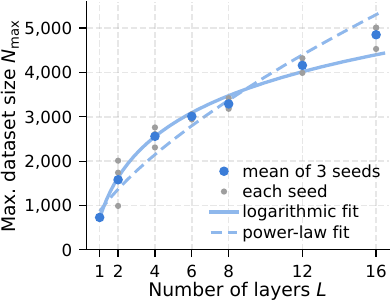}
    \end{minipage}\hfill
  \begin{minipage}[c]{0.49\linewidth}
    \caption{Maximum memorizable dataset size $N_{\max}$ vs.\ number of
    layers $L$ at fixed width $d=8$ (\texttt{Uniform} dataset). Grey dots:
    the three seeds; blue: their mean. Lines: the logarithmic (solid) and
    power-law (dashed) fits of Table~\ref{tab:depth-forms}, the two forms
    that the lack-of-fit test does not reject.}
    \label{fig:depth-expfit}
  \end{minipage}
\end{figure}

Figure~\ref{fig:depth-expfit} and Table~\ref{tab:depth-fits} (Appendix~\ref{sec:Appendix-results}) report the maximum memorizable dataset size $N_{\max}$ as a function of layer count $L$ at fixed width $d=8$ on \texttt{Uniform} data. At fixed width, the maximum memorizable dataset size grows with depth up to $L = 16$ layers, but sublinearly. Over the measured range a logarithmic form fits best, but the data do not distinguish between logarithmic and power-law scaling (Table~\ref{tab:depth-forms}, Appendix~\ref{sec:Appendix-results}).

 Two features of the experiment bear on this result. First, the training budget is the same at every depth: 40{,}000 iterations after warmup, with a hyperparameter search of 60 trials per learning-rate schedule. The reported capacity at each depth comes from three training runs, one per seed, each at the largest dataset size at which the run does not fail (Section~\ref{sec:dim-search}). The median iteration at which they converge is 51\% of the budget at $L = 1$ and between 77\% and 91\% at $L \geq 2$. This suggests that, at larger values of $L$, a larger budget can increase the measured capacity, which we leave to future work. It's also interesting to note that training is also less stable at larger depth. The fraction of training runs that reach 100\% accuracy but do not hold it for 50 consecutive steps rises from 2.9\% at $L = 1$ to 10.3\% at $L = 8$. Second, memorization is a task that one layer can already do at a rate proportional to its parameter count: \citet{nichani2024understanding} prove that a single layer of self-attention followed by an MLP reaches 100\% accuracy on factual recall if the parameter count of either component is linear in the number of facts, up to log factors. \citet{chen2024what} find that one attention layer suffices for memorization while reasoning and generalization need at least two or three.

\subsection{Ablation Studies}
\label{sec:ablation}

\paragraph{Impact of suboptimal hyperparameters.} To test whether the training protocol alone can account for the difference in scaling with prior work, we ran the width experiment on the same three dataset families and the same architecture while restricting the hyperparameter search to the learning rate and the batch size, which is the extent of tuning in prior work~\citep{lu2024scaling, morris2026how}. Under this restricted search, a power-law fit $d_{\min} \propto N^{c}$ gives an exponent $c$ between 0.73 and 0.77 for each of the three families (Figure~\ref{fig:min-dim-vs-size-fit}, Appendix~\ref{sec:Appendix-results}), so $P = \Theta(d^2)$ grows as $N^{1.46}$ to $N^{1.54}$: the number of non-embedding parameters grow superlinearly in $N$, which is even less favorable scaling than the linear relationship reported by prior work. \texttt{Symmetric} and \texttt{2Hop} require the same dimension as \texttt{Uniform}. Expanding optimization to include weight decay, minimum learning rate, learning rate schedule, and decay duration transforms both findings: Capacity scales exponentially with respect to the transformer dimension for \texttt{Uniform} (Figure~\ref{fig:min-dim-consolidated}), and the model's ability to leverage structure emerges---structured datasets require fewer parameters than \texttt{Uniform} at every dataset size in $N \ge 360$.

\paragraph{Batching and maximum training batches.} Beyond hyperparameters, two factors proved important. We found that random batch sampling caused loss spikes from adverse batches, requiring many batches to recover and inflating our estimate of the minimum transformer dimension $d_{\min}$; deterministic full-batch updates eliminated this issue (Appendix~\ref{app:sampling}). Since each iteration uses the full dataset, each iteration is one exposure per fact, and we use a higher maximum number of exposures per fact (up to 40{,}000) than \citet{lu2024scaling}. This is necessary because convergence slows dramatically as the model dimension approaches $d_{\min}$ (Appendix E, Figure~\ref{fig:batches_vs_dim}). Full-batch training also keeps the number of exposures per fact the same for every dataset size, because every iteration uses every fact. With a fixed batch size $B$ and a fixed number of steps  $T$ \citep{morris2026how}, each fact is seen $BT/N$ times during training on average, which decreases as the dataset grows. %

\subsection{Generality across architectures: OLMo-3}
\label{sec:exp-olmo}

To test whether our findings are specific to the GPT-2 architecture, we repeated the width scaling experiment on the OLMo-3 architecture at 8 dataset sizes, $N = 45$, $90$, $180$, $360$, $780$, $1{,}560$, $3{,}120$, and $6{,}240$, under the same experimental protocol; the OLMo-3 training details are given in Appendix~\ref{app:olmo3-details}.
Over this range, OLMo-3 shows the same scaling behavior as GPT-2 (Table~\ref{tab:olmo-width}, Appendix~\ref{sec:Appendix-results}). The minimum transformer model dimension grows sublinearly with the dataset size, a lack-of-fit F-test rejects a linear fit ($p = 8.8\times10^{-5}$), and neither a logarithmic nor a power-law fit is rejected, as with GPT-2 at these sizes. OLMo-3 memorizes each dataset at a smaller model dimension than GPT-2.

\section{Analysis}
\label{sec:analysis}
Our experiments reveal that memorization capacity grows exponentially with transformer dimension, with structured relations reaching a given capacity at a smaller dimension than unstructured data. For the \texttt{Uniform} dataset, the best-fit form for $d_{\min}(N)$ is logarithmic-with-offset, $d = 2.756\,\log(N + 80.6) - 10.09$ ($R^2 = 0.997$); a lack-of-fit F-test rejects the power-law alternatives (Table~\ref{tab:width-fits}). Under this fit, non-embedding parameters $P = \Theta(d^2)$ grow as $(\log N)^2$, which is sublinear in $N$; equivalently, capacity grows superlinearly in $P$. Structured datasets (\texttt{Symmetric}, \texttt{2Hop}) achieve further improvements in efficiency. Over the measured range, the logarithmic fit for $d_{\min}$ implies $P \sim (\log N)^2$, which grows more slowly than $\sqrt{N}$, the theoretical worst-case bound proved by \citet{kajitsuka2025}, although within our range $P$ itself is larger than $\sqrt{N}$ (4{,}617 against 223 at $N = 49{,}920$).
 
\paragraph{Embedding versus non-embedding parameters.} Thus far, our investigation has focused on the effect of non-embedding parameters on memorization capacity. But how much do the embedding parameters contribute to the capacity? In the width experiments on \texttt{Uniform} ($L = 1$, $m = 3$ relations, $N = 45$ to $49{,}920$), the non-embedding parameters and the embedding table scale differently. The number of non-embedding parameters, $12d^2 + 15d$, grows quadratically in $d$. Following standard convention (Section~\ref{sec:methods}), the embedding table stores one $d$-dimensional vector per token, resulting in $n + m = N/m + m$ rows, a number that grows linearly in $N$. As $d_{\min}$ grows logarithmically as a function of $N$, the table grows \emph{exponentially} versus $d_{\min}$, and its share of all parameters increases from 25\% at $N = 45$ to 99\% at $N = 49{,}920$ (Table~\ref{tab:bits-split}, Appendix~\ref{sec:Appendix-results}). Counting every parameter, a dataset with $N \log_2 n$ bits yields 0.72 to 1.80 bits per parameter at the five smallest dataset sizes ($N \le 780$) and 2.04 to 2.19 at the six largest dataset sizes $N \ge 1{,}560$. Therefore, bits per parameter increases when $N$ is small and shows signs of saturation when $N$ is large. A lack-of-fit F-test rejects the two non-saturating forms, a logarithm ($p = 7.4\times10^{-5}$) and a power-law ($p = 1.9\times10^{-6}$), but not the saturating forms $a - b\,N^{-c}$ and $a\,N/(N+k)$ ($p = 0.99$ and $0.93$).

Prior work reports ${\sim}2$ bits per parameter for GPT-2, LLaMA and Mistral models \citep{allenzhu2024physics} and ${\sim}3.6$ for GPT-style models \citep{morris2026how}. In comparison, we measure 2.04 to 2.19 bits per parameter at six dataset sizes $N \ge 1{,}560$. Both prior studies use a fixed vocabulary, and \citet{allenzhu2024physics} utilize several GPT-2 tokens to encode each entity. A fixed vocabulary is a practical choice, but composing entities from several tokens could confound the measurement of memorization capacity in terms of the number of facts. However, if each entity corresponds to a vocabulary item, a fixed vocabulary would limit the maximum number of expressible facts and would induce an artificial ceiling on the measured information capacity. %
Therefore, in our above experiments, we allowed the vocabulary to vary as a function of dataset size, where each entity and relation in the dataset is encoded as a unique token. %
But this raises the question as to how $d_{\text{min}}$ varies as a function of vocabulary size. To measure this, we conduct an additional experiment where we hold $N = 9{,}216$ fixed and vary the dataset from 9{,}216 entities with 1 relation to 96 entities with 96 relations, thereby varying the vocabulary size by a factor of 48 (Appendix~\ref{sec:exp-sweep}). Every dataset remains memorizable, and $d_{\min}$ rises from 6 ($m = 1$) to 46.33 ($m = 64$) before falling to 45.0 ($m = 96$). %
We note that this experiment does not test whether $d_{\min}$ still grows logarithmically in $N$ under an embedding table of fixed size (and sufficient capacity).

\section{Conclusion and Future Work}
 
We systematically measured transformer memorization capacity along width and depth axes and find that, under fixed budget, capacity scales exponentially with transformer dimension, and sublinearly with transformer depth. The minimum transformer dimension $d_{\min}$ required to memorize $N$ facts grows logarithmically in $N$, in contrast to the linear $P \sim N$ scaling reported by prior empirical studies. Our results further shed light on aspects of scaling law for memorization capacity that were missing in prior work.

\paragraph{Limitations.} Our experiments span $N \in [45, 49{,}920]$. Whether the logarithmic law for $d_{\min}(N)$ extends beyond the measured range is an empirical question. Our measurements use synthetic triples whose information content is exactly known; testing capacity on natural-language datasets such as $\texttt{bioS}$ \citep{allenzhu2024physics}, where the model must both store the facts and learn to parse and express them in natural language, is an interesting next step.

\paragraph{Future work.}

Our scaling experiments vary one axis at a time. Measuring capacity over the joint space of width and depth would help to further illuminate scaling behavior and identify potential trade-offs between the axes; we leave this to future work.

\label{end-of-main-text}
\subsection*{AI use statement}
In this work, we used generative AI tools (large language models) for writing and editing the text of the paper, and for writing research code. We have reviewed all AI-assisted work; every number in the paper was recomputed from the experiment databases by scripts that the authors inspected, every cited paper was read and its bibliographic entry checked by an author, and AI-written code was audited and tested by the authors before use. We take responsibility for the final content of this work, including text, claims or artifacts produced with the aid of generative AI.

\subsection*{Ethics statement}
All datasets in this work are synthetic (subject, relation, object) triples generated by the procedures in Appendix~\ref{sec:Appendix-algorithms}. The work involves no human subjects, personal data, or scraped text, and we are not aware of potential conflicts of interest.

\subsection*{Reproducibility statement}
The dataset-generation procedures are given in Algorithms~\ref{alg:gen-dataset} and~\ref{alg:powerlaw-batch}; the model and the training protocol in Section~\ref{sec:methods}; the hyperparameter search space, the search procedures, and the search objectives in Appendix~\ref{app:hyperopt}; and the statistical test used to compare scaling forms in Appendix~\ref{app:loft}. The code to reproduce the experiments and the results is open source and publicly available at \url{https://github.com/Magronox/transformer_information_capacity}.

\subsection*{Acknowledgments}

We thank the anonymous reviewers for their considerate and helpful feedback. This work was supported in part by the Rosen Center for Advanced Computing (RCAC) resources, services, and staff expertise at Purdue University.

\label{start-of-references}

\bibliography{\iclrdir iclr2027_refs}

\begin{thebibliography}{33}
\providecommand{\natexlab}[1]{#1}
\providecommand{\url}[1]{\texttt{#1}}
\expandafter\ifx\csname urlstyle\endcsname\relax
  \providecommand{\doi}[1]{doi: #1}\else
  \providecommand{\doi}{doi: \begingroup \urlstyle{rm}\Url}\fi

\bibitem[Akiba et~al.(2019)Akiba, Sano, Yanase, Ohta, and
  Koyama]{akiba2019optuna}
Takuya Akiba, Shotaro Sano, Toshihiko Yanase, Takeru Ohta, and Masanori Koyama.
\newblock Optuna: A next-generation hyperparameter optimization framework.
\newblock In \emph{Proceedings of the 25th ACM SIGKDD International Conference
  on Knowledge Discovery \& Data Mining}, KDD '19, pp.\  2623–2631, New York,
  NY, USA, 2019. Association for Computing Machinery.
\newblock ISBN 9781450362016.
\newblock \doi{10.1145/3292500.3330701}.
\newblock URL \url{https://doi.org/10.1145/3292500.3330701}.

\bibitem[Allen-Zhu \& Li(2025)Allen-Zhu and Li]{allenzhu2024physics}
Zeyuan Allen-Zhu and Yuanzhi Li.
\newblock Physics of language models: Part 3.3, knowledge capacity scaling
  laws.
\newblock In \emph{The Thirteenth International Conference on Learning
  Representations}, 2025.
\newblock URL \url{https://openreview.net/forum?id=FxNNiUgtfa}.

\bibitem[Bergstra et~al.(2011)Bergstra, Bardenet, Bengio, and
  K\'{e}gl]{NIPS2011_86e8f7ab}
James Bergstra, R\'{e}mi Bardenet, Yoshua Bengio, and Bal\'{a}zs K\'{e}gl.
\newblock Algorithms for hyper-parameter optimization.
\newblock In J.~Shawe-Taylor, R.~Zemel, P.~Bartlett, F.~Pereira, and K.Q.
  Weinberger (eds.), \emph{Advances in Neural Information Processing Systems},
  volume~24. Curran Associates, Inc., 2011.
\newblock URL
  \url{https://proceedings.neurips.cc/paper_files/paper/2011/file/86e8f7ab32cfd12577bc2619bc635690-Paper.pdf}.

\bibitem[Bordes et~al.(2013)Bordes, Usunier, Garcia-Duran, Weston, and
  Yakhnenko]{bordes2013translating}
Antoine Bordes, Nicolas Usunier, Alberto Garcia-Duran, Jason Weston, and Oksana
  Yakhnenko.
\newblock Translating embeddings for modeling multi-relational data.
\newblock In \emph{Advances in Neural Information Processing Systems},
  volume~26, pp.\  2787--2795, 2013.

\bibitem[Carlini et~al.(2021)Carlini, Tram{\`e}r, Wallace, Jagielski,
  Herbert-Voss, Lee, Roberts, Brown, Song, Erlingsson, Oprea, and
  Raffel]{carlini2021extracting}
Nicholas Carlini, Florian Tram{\`e}r, Eric Wallace, Matthew Jagielski, Ariel
  Herbert-Voss, Katherine Lee, Adam Roberts, Tom Brown, Dawn Song, {\'U}lfar
  Erlingsson, Alina Oprea, and Colin Raffel.
\newblock Extracting training data from large language models.
\newblock In \emph{30th USENIX Security Symposium}, 2021.

\bibitem[Carlini et~al.(2023)Carlini, Ippolito, Jagielski, Lee, Tram{\`e}r, and
  Zhang]{carlini2023quantifying}
Nicholas Carlini, Daphne Ippolito, Matthew Jagielski, Katherine Lee, Florian
  Tram{\`e}r, and Chiyuan Zhang.
\newblock Quantifying memorization across neural language models.
\newblock In \emph{ICLR}, 2023.

\bibitem[Chen \& Zou(2024)Chen and Zou]{chen2024what}
Xingwu Chen and Difan Zou.
\newblock What can transformer learn with varying depth? case studies on
  sequence learning tasks.
\newblock In \emph{Forty-first International Conference on Machine Learning},
  2024.
\newblock URL \url{https://openreview.net/forum?id=YNbCbcGyXE}.

\bibitem[Dai et~al.(2022)Dai, Dong, Hao, Sui, Chang, and Wei]{dai2022knowledge}
Damai Dai, Li~Dong, Yaru Hao, Zhifang Sui, Baobao Chang, and Furu Wei.
\newblock Knowledge neurons in pretrained transformers.
\newblock In \emph{Proceedings of the 60th Annual Meeting of the Association
  for Computational Linguistics (Volume 1: Long Papers)}, pp.\  8493--8502,
  Dublin, Ireland, 2022. Association for Computational Linguistics.
\newblock \doi{10.18653/v1/2022.acl-long.581}.

\bibitem[Draper \& Smith(1998)Draper and Smith]{lof}
Norman~R. Draper and Harry Smith.
\newblock \emph{Applied Regression Analysis}.
\newblock Wiley, April 1998.
\newblock ISBN 9781118625590.
\newblock \doi{10.1002/9781118625590}.
\newblock URL \url{http://dx.doi.org/10.1002/9781118625590}.

\bibitem[Geva et~al.(2021)Geva, Schuster, Berant, and
  Levy]{geva2021transformer}
Mor Geva, Roei Schuster, Jonathan Berant, and Omer Levy.
\newblock Transformer feed-forward layers are key-value memories.
\newblock In \emph{Proceedings of the 2021 Conference on Empirical Methods in
  Natural Language Processing}, pp.\  5484--5495, Online and Punta Cana,
  Dominican Republic, 2021. Association for Computational Linguistics.
\newblock \doi{10.18653/v1/2021.emnlp-main.446}.

\bibitem[Henighan et~al.(2020)Henighan, Kaplan, Katz, Chen, Hesse, Jackson,
  Jun, Brown, Dhariwal, Gray, Hallacy, Mann, Radford, Ramesh, Ryder, Ziegler,
  Schulman, Amodei, and McCandlish]{henighan2020scaling}
Tom Henighan, Jared Kaplan, Mor Katz, Mark Chen, Christopher Hesse, Jacob
  Jackson, Heewoo Jun, Tom~B. Brown, Prafulla Dhariwal, Scott Gray, Chris
  Hallacy, Benjamin Mann, Alec Radford, Aditya Ramesh, Nick Ryder, Daniel~M.
  Ziegler, John Schulman, Dario Amodei, and Sam McCandlish.
\newblock Scaling laws for autoregressive generative modeling.
\newblock \emph{arXiv preprint arXiv:2010.14701}, 2020.

\bibitem[Hernandez et~al.(2024)Hernandez, Sharma, Haklay, Meng, Wattenberg,
  Andreas, Belinkov, and Bau]{hernandez2024lineardecodingm}
Evan Hernandez, Arnab~Sen Sharma, Tal Haklay, Kevin Meng, Martin Wattenberg,
  Jacob Andreas, Yonatan Belinkov, and David Bau.
\newblock Linearity of relation decoding in transformer language models.
\newblock In \emph{The Twelfth International Conference on Learning
  Representations}, 2024.
\newblock URL \url{https://openreview.net/forum?id=w7LU2s14kE}.

\bibitem[Hoffmann et~al.(2022)Hoffmann, Borgeaud, Mensch, Buchatskaya, Cai,
  Rutherford, de~Las~Casas, Hendricks, Welbl, Clark, Hennigan, Noland,
  Millican, van~den Driessche, Damoc, Guy, Osindero, Simonyan, Elsen, Rae,
  Vinyals, and Sifre]{hoffmann2022chinchilla}
Jordan Hoffmann, Sebastian Borgeaud, Arthur Mensch, Elena Buchatskaya, Trevor
  Cai, Eliza Rutherford, Diego de~Las~Casas, Lisa~Anne Hendricks, Johannes
  Welbl, Aidan Clark, Tom Hennigan, Eric Noland, Katie Millican, George van~den
  Driessche, Bogdan Damoc, Aurelia Guy, Simon Osindero, Karen Simonyan, Erich
  Elsen, Jack~W. Rae, Oriol Vinyals, and Laurent Sifre.
\newblock Training compute-optimal large language models.
\newblock In \emph{Proceedings of the 36th International Conference on Neural
  Information Processing Systems}, NeurIPS'22, 2022.

\bibitem[Johnson \& Lindenstrauss(1984)Johnson and
  Lindenstrauss]{johnson1984extensions}
William~B. Johnson and Joram Lindenstrauss.
\newblock Extensions of {L}ipschitz mappings into a {H}ilbert space.
\newblock In \emph{Conference in Modern Analysis and Probability}, volume~26 of
  \emph{Contemporary Mathematics}, pp.\  189--206. American Mathematical
  Society, Providence, RI, 1984.

\bibitem[Kajitsuka \& Sato(2024)Kajitsuka and Sato]{kajitsuka2024}
Takumi Kajitsuka and Issei Sato.
\newblock Are transformers with one-layer self-attention universal
  approximators?
\newblock In \emph{Proceedings of the 12th International Conference on Learning
  Representations (ICLR)}, 2024.
\newblock Single-layer $\Theta(d n N)$ upper bound.

\bibitem[Kajitsuka \& Sato(2025)Kajitsuka and Sato]{kajitsuka2025}
Takumi Kajitsuka and Issei Sato.
\newblock On the optimal memorization capacity of transformers.
\newblock In \emph{Proceedings of the 13th International Conference on Learning
  Representations (ICLR)}, 2025.
\newblock Tight $\Theta(\sqrt{N})$ bounds.

\bibitem[Kaplan et~al.(2020)Kaplan, McCandlish, Henighan, Brown, Chess, Child,
  Gray, Radford, Wu, and Amodei]{kaplan2020scaling}
Jared Kaplan, Sam McCandlish, Tom Henighan, Tom~B. Brown, Benjamin Chess, Rewon
  Child, Scott Gray, Alec Radford, Jeffrey Wu, and Dario Amodei.
\newblock Scaling laws for neural language models.
\newblock \emph{arXiv preprint arXiv:2001.08361}, 2020.

\bibitem[Kim et~al.(2023)Kim, Kim, and Mozafari]{kim2023provable}
Junghwan Kim, Michelle Kim, and Barzan Mozafari.
\newblock Provable memorization capacity of transformers.
\newblock In \emph{The Eleventh International Conference on Learning
  Representations}, 2023.
\newblock URL \url{https://openreview.net/forum?id=8JCg5xJCTPR}.

\bibitem[Kingma \& Ba(2015)Kingma and Ba]{kingma2015adam}
Diederik~P. Kingma and Jimmy Ba.
\newblock Adam: A method for stochastic optimization.
\newblock In \emph{International Conference on Learning Representations
  (ICLR)}, 2015.
\newblock URL \url{https://arxiv.org/abs/1412.6980}.

\bibitem[Li et~al.(2025)Li, Kudugunta, and Zettlemoyer]{li2025misfitting}
Margaret Li, Sneha Kudugunta, and Luke Zettlemoyer.
\newblock (mis)fitting scaling laws: A survey of scaling law fitting techniques
  in deep learning.
\newblock In \emph{The Thirteenth International Conference on Learning
  Representations}, 2025.
\newblock URL \url{https://openreview.net/forum?id=xI71dsS3o4}.

\bibitem[Lu et~al.(2024)Lu, Li, Cheng, Ding, Huang, and Qiu]{lu2024scaling}
Xingyu Lu, Xiaonan Li, Qinyuan Cheng, Kai Ding, Xuanjing Huang, and Xipeng Qiu.
\newblock Scaling laws for fact memorization of large language models.
\newblock \emph{arXiv preprint arXiv:2406.15720}, 2024.

\bibitem[Lucibello \& M{\'e}zard(2024)Lucibello and
  M{\'e}zard]{lucibello2024exponential}
Carlo Lucibello and Marc M{\'e}zard.
\newblock The exponential capacity of dense associative memories.
\newblock \emph{Physical Review Letters}, 132\penalty0 (7):\penalty0 077301,
  2024.

\bibitem[Madden et~al.(2025)Madden, Fox, and Thrampoulidis]{madden2024}
Liam Madden, Curtis Fox, and Christos Thrampoulidis.
\newblock Next-token prediction capacity: General upper bounds and a lower
  bound for transformers.
\newblock \emph{IEEE Transactions on Information Theory}, 71\penalty0
  (9):\penalty0 7134--7148, 2025.
\newblock \doi{10.1109/TIT.2025.3584013}.

\bibitem[Mahdavi et~al.(2024)Mahdavi, Liao, and
  Thrampoulidis]{mahdavi2023memorization}
Sadegh Mahdavi, Renjie Liao, and Christos Thrampoulidis.
\newblock Memorization capacity of multi-head attention in transformers.
\newblock In \emph{Proceedings of the 12th International Conference on Learning
  Representations (ICLR)}, 2024.
\newblock Originally arXiv:2306.02010, 2023.

\bibitem[Meng et~al.(2022)Meng, Bau, Andonian, and Belinkov]{meng2022locating}
Kevin Meng, David Bau, Alex Andonian, and Yonatan Belinkov.
\newblock Locating and editing factual associations in {GPT}.
\newblock In \emph{Advances in Neural Information Processing Systems},
  volume~35, pp.\  17359--17372, 2022.

\bibitem[Morris et~al.(2026)Morris, Sitawarin, Kokhlikyan, Guo, Suh, Rush,
  Chaudhuri, and Mahloujifar]{morris2026how}
John~Xavier Morris, Chawin Sitawarin, Narine Kokhlikyan, Chuan Guo, G.~Edward
  Suh, Alexander~M Rush, Kamalika Chaudhuri, and Saeed Mahloujifar.
\newblock How much can language models memorize?, 2026.
\newblock URL \url{https://openreview.net/forum?id=NhU661EZ9C}.

\bibitem[Nichani et~al.(2025)Nichani, Lee, and
  Bietti]{nichani2024understanding}
Eshaan Nichani, Jason~D. Lee, and Alberto Bietti.
\newblock Understanding factual recall in transformers via associative
  memories.
\newblock In \emph{The Thirteenth International Conference on Learning
  Representations}, 2025.
\newblock URL \url{https://openreview.net/forum?id=hwSmPOAmhk}.

\bibitem[Petroni et~al.(2019)Petroni, Rockt{\"a}schel, Riedel, Lewis, Bakhtin,
  Wu, and Miller]{petroni2019language}
Fabio Petroni, Tim Rockt{\"a}schel, Sebastian Riedel, Patrick Lewis, Anton
  Bakhtin, Yuxiang Wu, and Alexander Miller.
\newblock Language models as knowledge bases?
\newblock In \emph{EMNLP-IJCNLP}, 2019.

\bibitem[Radford et~al.(2019)Radford, Wu, Child, Luan, Amodei, Sutskever,
  et~al.]{radford2019language}
Alec Radford, Jeffrey Wu, Rewon Child, David Luan, Dario Amodei, Ilya
  Sutskever, et~al.
\newblock Language models are unsupervised multitask learners.
\newblock \emph{OpenAI blog}, 1\penalty0 (8):\penalty0 9, 2019.

\bibitem[Ramsauer et~al.(2021)Ramsauer, Sch{\"a}fl, Lehner, Seidl, Widrich,
  Gruber, Holzleitner, Pavlovi{\'c}, Sandve, Greiff,
  et~al.]{ramsauer2020hopfield}
Hubert Ramsauer, Bernhard Sch{\"a}fl, Johannes Lehner, Philipp Seidl, Michael
  Widrich, Lukas Gruber, Markus Holzleitner, Milena Pavlovi{\'c}, Geir~Kjetil
  Sandve, Victor Greiff, et~al.
\newblock Hopfield networks is all you need.
\newblock In \emph{International Conference on Learning Representations}, 2021.

\bibitem[Ravfogel et~al.(2026)Ravfogel, Yehudai, Bruna, and
  Bietti]{ravfogel2026geometric}
Shauli Ravfogel, Gilad Yehudai, Joan Bruna, and Alberto Bietti.
\newblock Geometric factual recall in transformers, 2026.
\newblock URL \url{https://arxiv.org/abs/2605.12426}.

\bibitem[Tirumala et~al.(2022)Tirumala, Markosyan, Zettlemoyer, and
  Aghajanyan]{tirumala2022memorization}
Kushal Tirumala, Aram Markosyan, Luke Zettlemoyer, and Armen Aghajanyan.
\newblock Memorization without overfitting: Analyzing the training dynamics of
  large language models.
\newblock In \emph{Advances in Neural Information Processing Systems},
  volume~35, pp.\  38274--38290, 2022.

\bibitem[Yun et~al.(2020)Yun, Bhojanapalli, Rawat, Reddi, and
  Kumar]{DBLP:conf/iclr/YunBRRK20}
Chulhee Yun, Srinadh Bhojanapalli, Ankit~Singh Rawat, Sashank~J. Reddi, and
  Sanjiv Kumar.
\newblock Are transformers universal approximators of sequence-to-sequence
  functions?
\newblock In \emph{8th International Conference on Learning Representations,
  {ICLR} 2020, Addis Ababa, Ethiopia, April 26-30, 2020}. OpenReview.net, 2020.
\newblock URL \url{https://openreview.net/forum?id=ByxRM0Ntvr}.

\end{thebibliography}
\bibliographystyle{\iclrdir iclr2027_conference}

\newpage
\appendix
\makeatletter
\setlength{\intextsep}{6pt plus 2pt minus 2pt}\setlength{\textfloatsep}{8pt plus 2pt minus 2pt}\setlength{\floatsep}{6pt plus 2pt minus 2pt}
\def\section{\@startsection{section}{1}{\z@}{-1.2ex plus -0.3ex minus -.2ex}{0.8ex plus 0.2ex minus 0.2ex}{\large\sc\raggedright}}
\def\subsection{\@startsection{subsection}{2}{\z@}{-1.0ex plus -0.3ex minus -.2ex}{0.5ex plus .2ex}{\normalsize\sc\raggedright}}
\setlength{\@fptop}{0pt}\setlength{\@fpsep}{12pt plus 2pt}\setlength{\@fpbot}{0pt plus 1fil}
\makeatother
\renewcommand{\topfraction}{0.95}\renewcommand{\bottomfraction}{0.9}
\renewcommand{\textfraction}{0.05}\renewcommand{\floatpagefraction}{0.8}
\setcounter{topnumber}{4}\setcounter{bottomnumber}{4}\setcounter{totalnumber}{8}
\renewcommand{\theHfigure}{appendix.\arabic{figure}}
\renewcommand{\theHtable}{appendix.\arabic{table}}
\renewcommand{\theHalgorithm}{appendix.\arabic{algorithm}}
\section{Detailed Optimization Formulation}
\label{app:optimization}

We measure transformer memorization capacity in two complementary ways: in width experiments, we find the minimum transformer dimension $d^*$ for a fixed dataset; in depth experiments, we find the maximum dataset size $N^*$ for a fixed transformer. Both depend critically on training hyperparameters: suboptimal learning rates, weight decay, or schedules can prevent memorization even when the model has sufficient capacity.

Let $\theta$ denote the hyperparameter vector listed in Table~\ref{tab:hparam-search}. Let $\text{Train}(d, L, N; \theta)$ denote training a transformer of dimension $d$ and depth $L$ on a dataset of size $N$ with hyperparameters $\theta$ until either reaching 100\% accuracy or exhausting the training budget.

\noindent\textbf{Width experiments ($L = 1$, dataset size $N$ fixed).}
\begin{equation}
d^*(\theta; N) = \min\{\, d \in \mathbb{N} : \text{Train}(d, 1, N; \theta) \text{ achieves 100\% accuracy}\,\}.
\label{eq:width-obj}
\end{equation}
We seek $\theta^*(N) = \operatorname*{argmin}_\theta d^*(\theta; N)$ for each $N$ in our experimental sweep.

\noindent\textbf{Depth experiments (dimension $d$ and number of layers $L$ fixed).}
\begin{equation}
N^*(\theta; d, L) = \max\{\, N \in \mathbb{N} : \text{Train}(d, L, N; \theta) \text{ achieves 100\% accuracy}\,\}.
\label{eq:depth-obj}
\end{equation}
We seek $\theta^*(d, L) = \operatorname*{argmax}_\theta N^*(\theta; d, L)$ for each $L$.

Procedural details (per-trial search, multi-seed evaluation, BO objective formula) are in Appendix~\ref{app:hyperopt}.
 
\section{Hyperparameter Search Procedure}
\label{app:hyperopt}

\begin{table}[H]
\centering
\caption{Hyperparameter search space. Decay schedule is treated as a stratification variable: a separate optimization study is run for each schedule type and the best is reported. Warmup is fixed (not searched), as initial optimization results showed this would not hurt the best value achieved. Discussed in Section~\ref{sec:methods}.}
\vspace{4pt}
\label{tab:hparam-search}
\small
\begin{tabular}{lll}
\toprule
\textbf{Parameter} & \textbf{Search space} & \textbf{Notes} \\
\midrule
Learning rate & $[10^{-6},\,1]$ (log) & depth: $[10^{-7},\,3]$ \\
Use weight decay & $\{\text{False},\,\text{True}\}$ & gate; if False, $\text{wd}=0$ \\
Weight decay & $[10^{-8},\,10^{-1}]$ (log) & only sampled when gate is True \\
Min-LR ratio & $[10^{-8},\,1.0]$ (log) & $\text{min\_lr} = \text{lr} \cdot \text{ratio}$ \\
\midrule
\multicolumn{3}{l}{\textit{Schedule-shape:}} \\
Decay end step & $[100,\,40{,}000]$ & step (after warmup) where decay ends \\
\midrule
LR decay schedule & $\{\texttt{linear},\,\texttt{cosine}\}$ & stratified (separate study per type) \\
Warmup steps & $500 \cdot L$ & fixed; depth: $\min(500 \cdot L,\, 4000)$ \\
\bottomrule
\end{tabular}
\end{table}
Note that log in the table means the optimization was done in the log space.

\noindent\textbf{Per-trial dimension search (width experiments, Algorithm~\ref{alg:min-dim}).}
For each candidate configuration $\theta$ and seed $s$, we locate $d^*_s(\theta)$ as follows:
\begin{enumerate}
\item \emph{Geometric halving downward.} Start at $d = 32$ and halve ($d = 32, 16, 8, \ldots$), training at each $d$ and recording the smallest passing dimension.
\item \emph{Upward fallback.} If $d = 32$ itself fails, double upward ($d = 64, 128, 256$) until a passing dimension is found. If none passes by $d = 256$, the seed is marked as failed.
\item \emph{Binary search.} Refine between the last passing and first failing dimension to pinpoint $d^*_s$.
\end{enumerate}
\noindent\textbf{Per-trial dataset-size search (depth experiments, Algorithm~\ref{alg:max-size}).}
For each candidate configuration $\theta$ and seed $s$, we locate $N^*_s(\theta)$ as follows:
\begin{enumerate}
\item \emph{Geometric doubling/halving.} Starting from an initial estimate of $N$, double or halve the dataset size until the first success--failure boundary is found.
\item \emph{Binary search.} Refine between the last passing and first failing dataset size to pinpoint $N^*_s$.
\end{enumerate}

\noindent\textbf{Multi-seed objective.}
Each candidate configuration is evaluated jointly on three seeds (10, 20, 30) run in parallel across available GPUs, and is scored by
\begin{equation}
\text{objective} = \min_i d_i + 0.1\, n_{\text{failed}} + 0.05 \sum_i \mathbf{1}\{d_i < \infty\}\,(d_i - \min_j d_j),
\end{equation}
where $d_i$ is the smallest transformer model dimension at which the training run with seed $i$ memorizes the dataset, $d_i = \infty$ if seed $i$ fails to memorize within the training budget, and $n_{\text{failed}} = \#\{i : d_i = \infty\}$ is the number of failed seeds. The objective primarily minimizes the transformer model dimension; the two remaining terms serve only as tiebreakers between hyperparameter configurations that reach the same minimum dimension, preferring configurations with more reliable training (fewer failed seeds and less spread across seeds). The coefficients 0.1 and 0.05 are chosen small enough that the tiebreakers cannot outweigh the minimum dimension: because the dimension is an integer, changing the winning configuration would require the tiebreaker terms to add up to at least 1, which in turn would require the seeds to disagree by about 20 dimensions in total. No completed configuration came close to this.

\noindent\textbf{Multi-seed objective (depth experiments).}
In the depth experiments, each seed runs its own dataset-size search, and each hyperparameter configuration is scored by
\begin{equation}
\text{objective} = \max_i N_i - 0.1\, n_{\text{failed}}\, S - 0.05 \sum_i \mathbf{1}\{N_i > 0\}\,(\max_j N_j - N_i),
\end{equation}
where $N_i$ is the largest dataset size that the training run with seed $i$ memorizes, $N_i = 0$ if seed $i$ memorizes no dataset size, $n_{\text{failed}} = \#\{i : N_i = 0\}$ is the number of failed seeds, and $S = 1{,}002$ facts is the coarse step of the size search. The objective primarily maximizes the dataset size; the two remaining terms prefer configurations with more reliable training (fewer failed seeds and less spread across seeds). Unlike in the width experiments, these terms are measured in facts and can change which configuration scores highest. The objective only guides the search: for each depth we report the configuration with the largest mean over the three seeds (Table~\ref{tab:depth-fits}), which does not use the objective.

\paragraph{Lack-of-fit F-test.}
\label{app:loft}
To test whether a functional form $f$ with $k$ parameters describes a scaling curve measured at $m$ points $x_1, \dots, x_m$ (dataset sizes, or numbers of layers), we use the lack-of-fit F-test for repeated measurements~\citep{lof}. At each point we have three values $y_{i1}, y_{i2}, y_{i3}$, one per seed, where $y$ is the logarithm of the measured quantity ($d_{\min}$, $N_{\max}$, or bits per parameter). Let $\bar{y}_i$ be their mean. We fit $\log f$ to the means $\bar{y}_i$ by least squares, and compute the pure-error and lack-of-fit sums of squares
\begin{equation*}
\mathrm{SS}_{\mathrm{PE}} = \sum_{i=1}^{m} \sum_{j=1}^{3} (y_{ij} - \bar{y}_i)^2, \qquad
\mathrm{SS}_{\mathrm{LOF}} = 3 \sum_{i=1}^{m} \bigl(\bar{y}_i - \log f(x_i)\bigr)^2,
\end{equation*}
with $2m$ and $m - k$ degrees of freedom. If $f$ is the correct form, $F = \frac{\mathrm{SS}_{\mathrm{LOF}}/(m-k)}{\mathrm{SS}_{\mathrm{PE}}/(2m)}$ follows an $F(m-k,\, 2m)$ distribution; we report $p = \Pr[F(m-k, 2m) > F]$ and reject the form when $p < 0.05$. The test asks whether the fitted curve misses the means by more than the spread across seeds explains, so a large $p$ means the form is consistent with the data, not that it is proven.

\subsection{Generality Across Architectures: OLMo-3 Implementation and Results}
\label{app:olmo3-details}

To evaluate whether logarithmic width scaling is specific to GPT-2 or generalizes across modern transformer architectures, we repeat the width scaling experiment using the OLMo-3 architecture. 

To maintain parameter fairness against our GPT-2 baseline (Section 3.2) at width $d$, the OLMo-3 block is aligned via three adjustments: 

\begin{enumerate}[leftmargin=*]
    \item Disabling Positional Embeddings for position-free processing across all integer widths $d$;
    \item Using standard full attention matching GPT-2; and
    \item Setting the SwiGLU (Swish-Gated Linear Unit) feed-forward hidden dimension to $\mathrm{round}(8d/3)$ so that its 3-matrix FFN parameter count matches GPT-2's standard 2-matrix FFN up to rounding.
\end{enumerate}

Models are trained using full-batch gradient descent, bfloat16 mixed precision, and the Adam optimizer. For each dataset size $N$, hyperparameters are optimized over the search space of Table~\ref{tab:hparam-search} with 60 Optuna TPE trials per learning-rate schedule. Convergence requires maintaining 100\% training accuracy for 50 consecutive evaluation steps within a budget of 40,000 batches.

As detailed in Table~\ref{tab:olmo3-results-detailed}, the minimum required dimension $d_{\min}$ for OLMo-3 grows from $d=2$ at $N=45$ to $d=6$ at $N=6{,}240$; as for GPT-2, a lack-of-fit F-test rejects a linear fit ($p = 8.8\times10^{-5}$) while a logarithm ($p = 0.68$) and a power law with exponent 0.22 ($p = 0.25$) both pass.

\begin{table}[H]
\centering
\caption{OLMo-3 factual capacity measurements across dataset sizes $N$. $d_{\min}$ is the minimum width achieving 100\% memorization accuracy across three random seeds (seeds 10, 20, 30).}
\label{tab:olmo3-results-detailed}
\vspace{4pt}
\small
\begin{tabular}{rccccc}
\toprule
\textbf{Entities ($n$)} & \textbf{Facts ($N$)} & \textbf{Information (Bits)} & \textbf{Seeds (10 / 20 / 30) $d_{\min}$} & \textbf{Mean $d_{\min}$} & \textbf{Best $d_{\min}$} \\
\midrule
15   & 45    & 176    & 2 / 2 / 2 & 2.00 & \textbf{2} \\
30   & 90    & 442    & 3 / 4 / 2 & 3.00 & \textbf{2} \\
60   & 180   & 1,063  & 3 / 3 / 3 & 3.00 & \textbf{3} \\
120  & 360   & 2,486  & 4 / 4 / 4 & 4.00 & \textbf{4} \\
260  & 780   & 6,257  & 4 / 4 / 5 & 4.33 & \textbf{4} \\
520  & 1,560 & 14,075 & 5 / 5 / 5 & 5.00 & \textbf{5} \\
1040 & 3,120 & 31,270 & 6 / 6 / 6 & 6.00 & \textbf{6} \\
2080 & 6,240 & 68,780 & 6 / 6 / 6 & 6.00 & \textbf{6} \\

\bottomrule
\end{tabular}
\end{table}

\section{Sampling Sensitivity Analysis}
\label{app:sampling}

\begin{figure}[!ht]
  \centering
  \begin{minipage}[c]{0.47\linewidth}
    \includegraphics[width=\linewidth]{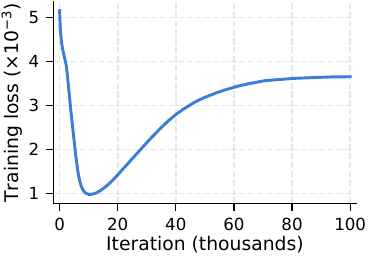}
  \end{minipage}\hfill
  \begin{minipage}[c]{0.49\linewidth}
  \caption{Training loss vs.\ number of iterations under random batch sampling (\texttt{Uniform} dataset of size 1500, transformer dimension 95). Sudden spikes correspond to ``bad batches''---unlucky random draws whose gradient is misaligned with the loss surface, causing transient divergence and slowing convergence. These episodes inflate measured $d^*$ when the iteration cap is finite. Switching to deterministic full-batch updates eliminates them and reduces the minimum dimension required for memorization.}
  \label{fig:loss-vs-max-iterations-uniform}
  \end{minipage}
\end{figure}

Throughout our experiments, we found that memorization—and, consequently, the estimated minimum transformer dimension $d^*$—was highly sensitive to \emph{random sampling} of training examples. This sensitivity persisted even when the sample size used per update exceeded the dataset size by factors (random sampling with replacement), indicating that the issue was not whether the sample was small or large, but rather the randomness of the draw itself. In particular, we observed that a single adverse sample (a ``bad batch'' in the sense of an unlucky random draw) could induce a large, misaligned update that slowed loss descent, as evidenced by spikes or plateaus in the training loss traces (Figure~\ref{fig:loss-vs-max-iterations-uniform}). When we removed this source of noise by using \emph{deterministic full-dataset updates} (i.e., the exact empirical gradient computed over a fixed example order), the minimum feasible dimension dropped significantly, and the region of settings that achieved this minimum became stable across seeds.
 
Based on this discovery, for all subsequent experiments, we fix the batch size: for \texttt{Uniform}, \texttt{Symmetric}, and \texttt{2Hop} datasets, we set $b = N$ (the full dataset size), processing all triples in each batch. For \texttt{PowerLaw} distributions, we use Power-Law sampling as mentioned in Appendix~\ref{app:powerlaw}.
 
\section{Power-Law Sampling}
\label{app:powerlaw}
 
To probe whether non-uniform fact distributions affect memorization capacity, we sample triples according to a Zipf distribution. For a dataset with $n$ entities and $m=3$ relations ($nm$ total triples), each triple is assigned a rank $i = 1, 2, \ldots, nm$ and sampled with probability:
\begin{equation}
p_i = \frac{i^{-\alpha}}{Z}, \quad \text{where} \quad Z = \sum_{j=1}^{nm} j^{-\alpha}
\end{equation}
and we use $\alpha = 1.0$ in our experiments.
 
The replication rule is: each fact $i$ appears $\max(1, \lceil k\,p_i \rceil)$ times per batch, where the multiplier $k = 1/p_N$ is set so that the rarest fact (smallest $p_i$) appears exactly once per batch. The batch is constructed once and reused for every training iteration (Algorithm~\ref{alg:powerlaw-batch}). The resulting per-batch size grows roughly as $k$, which is why we capped tested $N$ at $12{,}480$ (Section~\ref{sec:datasets}).
 
\section{Additional Experimental Results}
\label{sec:Appendix-results}

\paragraph{Positional embeddings.} In preliminary experiments we trained models with and without learned positional embeddings. Since every input has the fixed form $(s, r)$, position is implicit in the input structure, and positional embeddings did not change the minimum transformer model dimension required for memorization; we therefore omit them in all experiments.
\begin{figure}[!ht]
  \centering
  \includegraphics[width=0.6\linewidth,trim={0 0 459 0},clip]{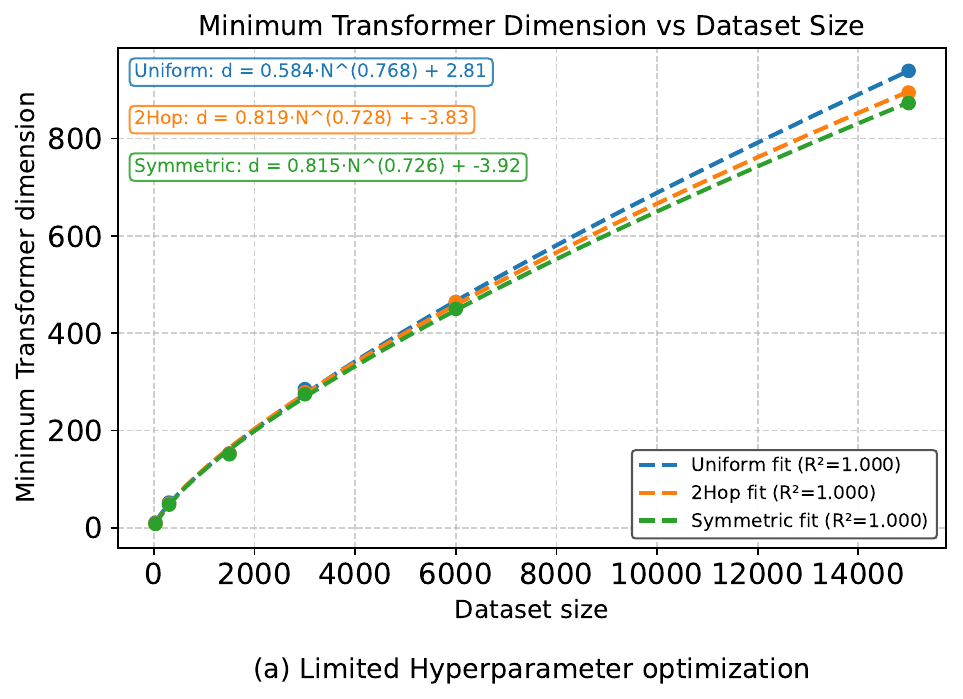}
  \caption{Minimum model dimension required for complete memorization vs.\ dataset size with power-law fit (offset unconstrained) across dataset types
 (\texttt{Uniform}, \texttt{Symmetric}, \texttt{2Hop}), when we only optimized learning rate and batch size. The minimum transformer dimension required for complete memorization is way higher than under the full search (Figure~\ref{fig:min-dim-consolidated}), highlighting the impact of hyperparameter optimization.}
  \label{fig:min-dim-vs-size-fit}
\end{figure}

\begin{table}[H]
\centering
\caption{Non-embedding-vs-embedding parameter breakdown for the smallest single-layer model that memorizes a \texttt{Uniform} dataset of size $N$ ($d_{\min}$ averaged over 3 seeds). Non-embedding: $P = 12 d^2 + 15d$; embedding: $P_{\text{emb}} = (E+R) \cdot d$; bits $= N \log_2 E$. Including embedding parameters, capacity rises from 0.72 to ${\approx}2.2$ bits per parameter, reaching ${\geq}2.0$ from $N = 1{,}560$ onward. Discussed in Section~\ref{sec:analysis}.}
\vspace{4pt}
\label{tab:bits-split}
\small
\setlength{\tabcolsep}{4pt}
\begin{tabular}{rrrrr}
\toprule
$N$ & $d_{\min}$ & Body \% & $\text{bits}/P_{\text{body}}$ & $\text{bits}/P_{\text{tot}}$ \\
\midrule
45        & 3.33  & 75\% & 1.0   & 0.72 \\
90        & 4.00  & 66\% & 1.8   & 1.15 \\
180       & 5.00  & 54\% & 2.8   & 1.54 \\
360       & 6.67  & 44\% & 3.9   & 1.71 \\
780       & 9.00  & 32\% & 5.7   & 1.80 \\
1{,}560   & 10.33 & 21\% & 9.8   & 2.06 \\
3{,}120   & 12.67 & 14\% & 14.8  & 2.04 \\
6{,}240   & 14.33 & 8\%  & 25.7  & 2.11 \\
12{,}480  & 16.00 & 5\%  & 45.3  & 2.15 \\
24{,}960  & 17.33 & 3\%  & 84.1  & 2.19 \\
49{,}920  & 19.00 & 1\%  & 151.6 & 2.18 \\
\bottomrule
\end{tabular}
\end{table}

\begin{table}[H]
\centering
\caption{Depth-scaling fits on \texttt{Uniform} (fixed width $d = 8$, 7 depths, 3 seeds). All fits are computed in log space; the last column is the lack-of-fit F-test $p$-value ($p < 0.05$ rejects the functional form).}
\vspace{4pt}
\label{tab:depth-forms}
\small
\begin{tabular}{llrr}
\toprule
\textbf{Functional form} & \textbf{Best fit} & $R^2$ & \textbf{Lack-of-fit $p$} \\
\midrule
Linear without intercept    & $N = 498\,L$                        & $0.690$ & $7.8 \times 10^{-6}$ \\
Linear with intercept       & $N = 338\,L + 597$                  & $0.906$ & $0.0025$ \\
Power-law, exponent $\geq 1$ & $N = 498\,L^{1.000}$               & $0.690$ & $4.0 \times 10^{-6}$ \\
Exponential                 & $N = 1197\,e^{0.103\,L}$            & $0.741$ & $1.2 \times 10^{-5}$ \\
\midrule
Logarithmic                 & $N = 1331\,\log L + 709$             & $0.992$ & $0.73$ \\
Power-law                   & $N = 870\,L^{0.652}$                & $0.966$ & $0.092$ \\
\bottomrule
\end{tabular}
\end{table}

\begin{table}[H]
\centering
\caption{OLMo-3 width scaling experiment: minimum transformer model dimension $d$ needed to memorize all $N$ facts, with $m = 3$ relations (averaged over 3 seeds; GPT-2 shown for comparison).}
\vspace{4pt}
\label{tab:olmo-width}
\small
\begin{tabular}{lrrrrrrrr}
\toprule
$N$ & 45 & 90 & 180 & 360 & 780 & 1{,}560 & 3{,}120 & 6{,}240 \\
\midrule
OLMo-3 & 2.00 & 3.00 & 3.00 & 4.00 & 4.33 & 5.00 & 6.00 & 6.00 \\
GPT-2  & 3.33 & 4.00 & 5.00 & 6.67 & 9.00 & 10.33 & 12.67 & 14.33 \\
\bottomrule
\end{tabular}
\end{table}

\begin{table}[H]
\centering
\caption{Depth scaling experiment at transformer model dimension 8: maximum dataset size memorized completely, for each number of layers $L$ (\texttt{Uniform}, $m = 3$). Each entry is the mean over seeds 10, 20, 30 for the hyperparameter configuration with the largest mean; the range is the smallest and largest seed.}
\vspace{4pt}
\label{tab:depth-fits}
\small
\begin{tabular}{lrrrrrrr}
\toprule
layers $L$ & 1 & 2 & 4 & 6 & 8 & 12 & 16 \\
\midrule
max $N$ (mean) & 730     & 1{,}580 & 2{,}560 & 3{,}010 & 3{,}290 & 4{,}160 & 4{,}850 \\
smallest seed  & 720     & 990     & 2{,}310 & 2{,}940 & 3{,}180 & 3{,}990 & 4{,}530 \\
largest seed   & 750     & 2{,}010 & 2{,}760 & 3{,}060 & 3{,}420 & 4{,}320 & 5{,}010 \\
\bottomrule
\end{tabular}
\end{table}

\subsection{Width scaling measured by the maximum dataset size}
\label{sec:exp-maxn-width}

Section~\ref{sec:exp-width} measures width scaling by finding the smallest dimension $d_{\min}$ that memorizes each dataset size $N$. The same scaling can also be measured by finding the largest dataset size $N_{\max}$ that each dimension memorizes. To confirm that the two methods result in similar scaling behavior, we ran the maximum-dataset-size search (Algorithm~\ref{alg:max-size}) at $L = 1$
for transformer dimensions $d = 3$ to $16$, with seeds 10, 20, 30 and 60 completed trials per width. We only ran this with linear learning-rate schedule, and the fine step for search is
3 facts instead of 30, because the memorization capacities at small $d$ are below 100 facts.
Table~\ref{tab:maxn-width} compares the measured $N_{\max}$ with the $N$ at which the logarithmic-with-offset fit of
Table~\ref{tab:width-fits} predicts $d_{\min} = d$; they differ by $-21\%$ to $+35\%$.
Table~\ref{tab:maxn-width-fits} fits the same forms as Table~\ref{tab:width-fits}, with the same
test and constraints. As with the minimum-dimension search (Table~\ref{tab:width-fits}), the
logarithmic-with-offset form is the only form that is not rejected ($p = 0.085$); both power laws,
the plain logarithm and the linear form are rejected.

\begin{table}[H]
\centering
\caption{Largest memorizable dataset size $N_{\max}$ at $L = 1$ for each width $d$ (mean over seeds
10, 20, 30), and the $N$ predicted by the width scaling experiment's logarithmic-with-offset fit $d = 2.756\ln(N + 80.6) - 10.09$ of
Table~\ref{tab:width-fits}.}
\vspace{4pt}
\label{tab:maxn-width}
\small
\setlength{\tabcolsep}{4pt}
\begin{tabular}{lrrrrrrrrrr}
\toprule
width $d$ & 3 & 4 & 5 & 7 & 8 & 9 & 10 & 13 & 14 & 16 \\
\midrule
$N_{\max}$ (measured) & 34 & 89 & 214 & 452 & 704 & 918 & 1{,}584 & 4{,}433 & 7{,}379 & 10{,}129 \\
$N$ (Table~\ref{tab:width-fits} fit)      & 35 & 85 & 158 & 413 & 628 & 939 & 1{,}384 & 4{,}270 & 6{,}173 & 12{,}841 \\
difference           & $-3\%$ & $+4\%$ & $+35\%$ & $+10\%$ & $+12\%$ & $-2\%$ & $+14\%$ & $+4\%$ & $+20\%$ & $-21\%$ \\
\bottomrule
\end{tabular}
\end{table}

\begin{table}[H]
\centering
\caption{Fits to the $N_{\max}$ measurements of Table~\ref{tab:maxn-width}, written as $d = f(N)$
as in Table~\ref{tab:width-fits}. Fits are computed on $\ln N_{\max}$ with the three seeds; the
offsets are constrained so that the fitted $d$ is positive for every $N > 0$; the last column is the
lack-of-fit F-test $p$-value ($p < 0.05$ rejects the functional form).}
\vspace{4pt}
\label{tab:maxn-width-fits}
\small
\begin{tabular}{llrr}
\toprule
\textbf{Functional form} & \textbf{Best fit} & $R^2$ & \textbf{Lack-of-fit $p$} \\
\midrule
Logarithmic\,+\,offset & $d = 3.018\,\ln(N + 125.9) - 12.31$ & $0.996$ & $0.085$ \\
Logarithmic            & $d = 2.374\,\ln N - 6.93$           & $0.970$ & $4.6 \times 10^{-7}$ \\
Power-law\,+\,offset   & $d = 1.08\,N^{0.298} + 0.00$        & $0.989$ & $5.5 \times 10^{-4}$ \\
Power-law              & $d = 1.08\,N^{0.298}$               & $0.989$ & $9.4 \times 10^{-4}$ \\
Linear                 & $d = 4.66 \times 10^{-3}\, N + 2.87$ & $0.816$ & $3.1 \times 10^{-14}$ \\
\bottomrule
\end{tabular}
\end{table}

\subsection{Varying the number of relations at fixed dataset size}
\label{sec:exp-sweep}

To test whether the growing vocabulary drives our results, we hold the number of facts fixed at $N = 9{,}216$ and vary the composition of the dataset---from 9{,}216 entities with 1 relation to 96 entities with 96 relations---shrinking the vocabulary (and with it the embedding table) 48-fold. Every dataset remains fully memorizable, and the minimum transformer model dimension rises from 6.00 with $m = 1$ relation to 46.33 with $m = 64$ relations, then falls to 45.00 with $m = 96$. A joint fit with the width scaling experiment's dataset sizes gives $d \approx 2.58\log n + 12.65\log m - 18.71$ ($R^2 = 0.973$), where $n$ is the number of entities and $m$ the number of relations; a lack-of-fit F-test rejects this form ($p < 10^{-10}$).

\begin{table}[H]
\centering
\caption{Minimum transformer model dimension $d$ needed to memorize $N = 9{,}216$ facts as we vary the composition of the dataset, from 9{,}216 entities with 1 relation to 96 entities with 96 relations (averaged over 3 seeds).}
\vspace{4pt}
\label{tab:er-sweep}
\footnotesize
\setlength{\tabcolsep}{3.5pt}
\begin{tabular}{lrrrrrrrrr}
\toprule
entities $n$ & 9216 & 4608 & 3072 & 2304 & 1152 & 576 & 288 & 144 & 96 \\
relations $m$ & 1 & 2 & 3 & 4 & 8 & 16 & 32 & 64 & 96 \\
\midrule
min.\ dimension $d$ to memorize all facts & 6.00 & 10.67 & 15.33 & 19.00 & 29.33 & 38.00 & 44.00 & 46.33 & 45.00 \\
\bottomrule
\end{tabular}
\end{table}

\begin{figure}[!ht]
  \centering
  \includegraphics[width=\linewidth]{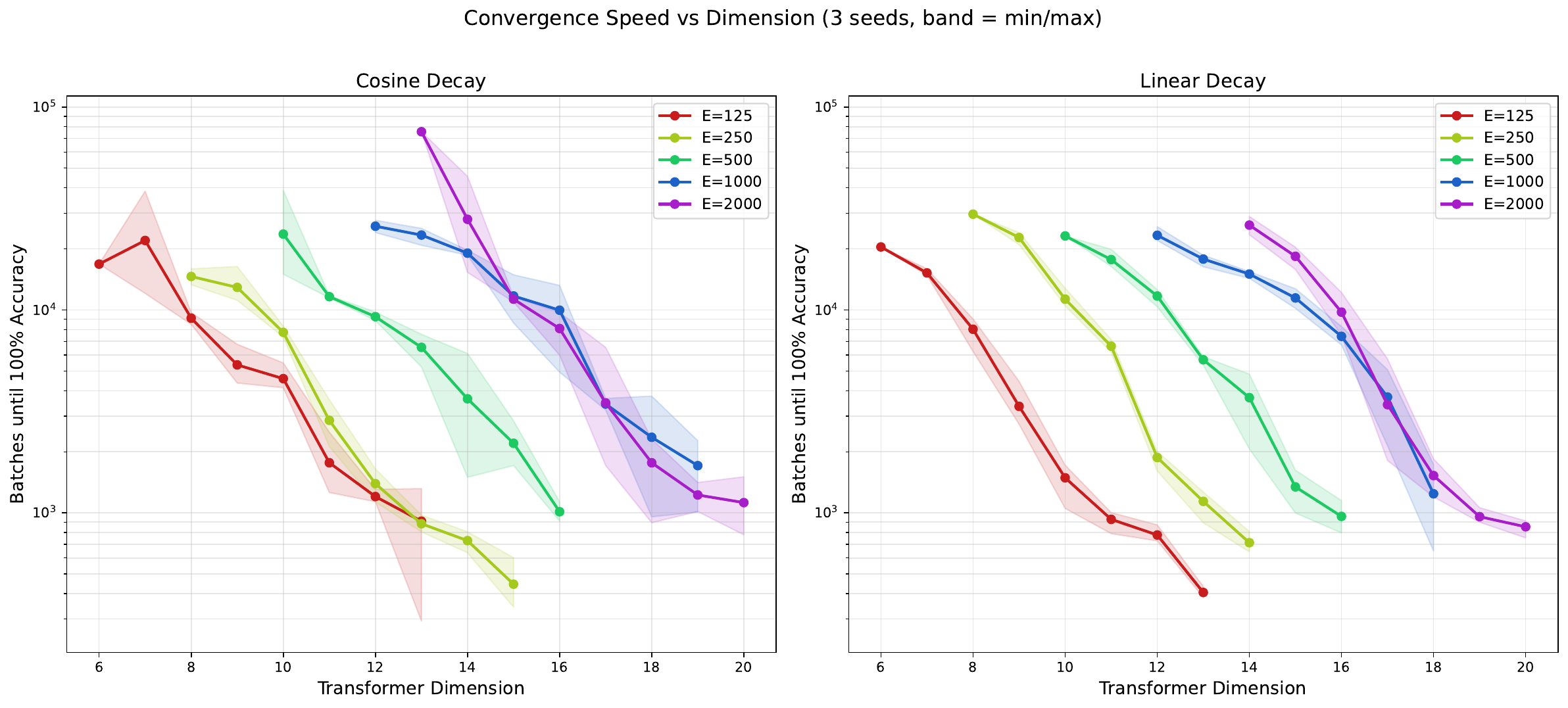}
  \caption{\textbf{Convergence Speed vs. Dimension.} The total number of  batches required to achieve 100\% training accuracy vs transformer dimension $d$ for different entity sizes $E \in \{125, 250, 500, 1000, 2000\}$. Plots for both the linear and cosine decay types are shown. The y-axis is on a log scale, showing that as the model approaches the theoretical minimum dimension $d_{min}$, the cost required to train for perfect memorization increases exponentially. The shaded bands represent the variation of the number of batches required over 3 seeds.}
  \label{fig:batches_vs_dim}
\end{figure}

Figure~\ref{fig:batches_vs_dim} illustrates the relationship between the number of batches required for $100\%$ training accuracy vs the transformer dimension $d$ across different dataset sizes. We observe a clear "capacity compute tradeoff": a smaller dimension requires many more training batches to achieve the same level of capacity as larger models.

Furthermore, the plot compares 2 different decay types and shows that both schedules eventually converge to a similar minimum dimension. This figure shows that a high number of batches is essential for uncovering the true scaling regime between transformer dimension and dataset size.

\section{Algorithms for Experiments}
\label{sec:Appendix-algorithms}
 
\begin{algorithm}[H]
\caption{Dataset generation: GenerateDataset($n$, $m$, $type$)}
\label{alg:gen-dataset}
\begin{algorithmic}[1]
\REQUIRE number of entities $n$, number of relations $m$, $type \in \{\texttt{Uniform}, \texttt{Symmetric}, \texttt{2Hop}\}$. Each entity and relation is a single vocabulary token ($|V| = n + m$). For \texttt{2Hop}, $m = 3$ and the last relation $r_3$ is dedicated to the two-hop compositions.
\STATE $D \gets \emptyset$
\FORALL{relations $r \in \{1, \ldots, m\}$}
\FORALL{subjects $s \in \{1, \ldots, n\}$}
  \IF{$type = \texttt{Uniform}$}
    \STATE sample $o$ uniformly from $\{1, \ldots, n\}$; add $(s, r, o)$ to $D$
  \ELSIF{$type = \texttt{Symmetric}$}
    \STATE add $(s, r, \pi_r(s))$ to $D$, where $\pi_r$ pairs consecutive entities of a uniformly random shuffle of the $n$ entities, drawn once for each relation $r$ (the last entity is paired with itself when $n$ is odd); $(\pi_r(s), r, s)$ is therefore in $D$ as well
  \ELSIF{$type = \texttt{2Hop}$ \AND $r \in \{r_1, r_2\}$}
    \STATE sample $o$ uniformly from $\{1, \ldots, n\}$; add $(s, r, o)$ to $D$
  \ELSIF{$type = \texttt{2Hop}$ \AND $r = r_3$}
    \STATE add $(s, r_3, c)$ to $D$, where $b$ and $c$ are the entities such that $(s, r_1, b) \in D$ and $(b, r_2, c) \in D$ (these facts exist because $r_3$ is processed last)
  \ENDIF
\ENDFOR
\ENDFOR
\STATE \textbf{return} $D$ \COMMENT{every (subject, relation) pair appears exactly once: $N = n \cdot m$ facts, each a 3-token sequence; the training loss is computed on the object token only}
\end{algorithmic}
\end{algorithm}

\begin{algorithm}[H]
\caption{Batch construction for \texttt{PowerLaw}: ConstructBatch($D$, $\alpha$)}
\label{alg:powerlaw-batch}
\begin{algorithmic}[1]
\REQUIRE dataset $D$ of $N$ facts, Zipf exponent $\alpha > 0$. The batch is constructed once and reused for every training iteration.
\STATE assign each fact $i \in D$ a Zipf probability $p_i \propto i^{-\alpha}$ (fixed for the dataset; they do not change across batches)
\STATE $k \gets 1/p_N$, where $p_N$ is the probability of the rarest fact \COMMENT{the rarest fact appears exactly once per batch}
\STATE construct the batch so that each fact $i$ appears $\max(1, \lceil k\, p_i \rceil)$ times
\STATE \textbf{return} the batch \COMMENT{for \texttt{Uniform}, \texttt{Symmetric}, and \texttt{2Hop}, each batch contains each fact exactly once}
\end{algorithmic}
\end{algorithm}

\begin{algorithm}[H]
\caption{Per-seed dimension search: MinDim($N$, $\theta$, $s$)}
\label{alg:min-dim}
\begin{algorithmic}[1]
\REQUIRE dataset size $N$, hyperparameters $\theta$, seed $s$; start $d_0 = 32$, cap $d_{\max} = 256$. \textsc{Train}$(d)$ returns pass if $\mathrm{Train}(d, 1, N; \theta)$ with seed $s$ reaches stable 100\% accuracy.
\ENSURE $d^*_s(\theta)$, or $\infty$ if no $d \leq d_{\max}$ passes
\STATE $d \gets d_0$; $d_{\text{ok}} \gets \infty$
\WHILE{$d \geq 1$ \textbf{and} \textsc{Train}$(d)$ passes}
  \STATE $d_{\text{ok}} \gets d$; $d \gets \lfloor d/2 \rfloor$ \COMMENT{geometric halving}
\ENDWHILE
\IF{$d_{\text{ok}} = \infty$}
  \STATE $d \gets 2 d_0$
  \WHILE{$d \leq d_{\max}$ \textbf{and} \textsc{Train}$(d)$ fails}
    \STATE $d \gets 2d$ \COMMENT{upward fallback}
  \ENDWHILE
  \IF{$d > d_{\max}$} \STATE \textbf{return} $\infty$ \ENDIF
  \STATE $d_{\text{ok}} \gets d$; $d_{\text{fail}} \gets d/2$
\ELSE
  \IF{$d = 0$} \STATE \textbf{return} $d_{\text{ok}}$ \COMMENT{every $d \geq 1$ passed} \ENDIF \STATE $d_{\text{fail}} \gets d$ \COMMENT{first failing dimension}
\ENDIF
\STATE $lo \gets d_{\text{fail}}$ ($d_{\text{fail}} + 1$ after the upward fallback); $hi \gets d_{\text{ok}}$ \COMMENT{both ends included}
\WHILE{$lo \leq hi$}
  \STATE $m \gets \lfloor (lo + hi)/2 \rfloor$
  \IF{\textsc{Train}$(m)$ passes} \STATE $d_{\text{ok}} \gets m$; $hi \gets m - 1$ \ELSE \STATE $lo \gets m + 1$ \ENDIF
\ENDWHILE
\STATE \textbf{return} $d_{\text{ok}}$
\end{algorithmic}
\end{algorithm}

\begin{algorithm}[H]
\caption{Per-seed dataset-size search: MaxSize($d$, $L$, $\theta$, $s$)}
\label{alg:max-size}
\begin{algorithmic}[1]
\REQUIRE width $d$, depth $L$, hyperparameters $\theta$, seed $s$; initial estimate $N_0$; coarse step $S = 1{,}002$ facts; fine step $F = 30$ facts. \textsc{Train}$(N)$ returns pass if $\mathrm{Train}(d, L, N; \theta)$ with seed $s$ reaches stable 100\% accuracy. $\mathrm{round}_S(\cdot)$ rounds to the nearest multiple of $S$, never below $S$; $\mathrm{round}_F(\cdot)$ likewise for $F$.
\ENSURE $N^*_s(\theta)$, or $0$ if no size passes
\IF{\textsc{Train}$(N_0)$ passes}
  \STATE $N_{\text{ok}} \gets N_0$; $N \gets \mathrm{round}_S(2 N_0)$
  \WHILE{\textsc{Train}$(N)$ passes} \STATE $N_{\text{ok}} \gets N$; $N \gets \mathrm{round}_S(2N)$ \ENDWHILE \COMMENT{doubling}
  \STATE $N_{\text{fail}} \gets N$
\ELSE
  \STATE $N_{\text{fail}} \gets N_0$; $N \gets \mathrm{round}_S(N_0 / 2)$
  \WHILE{\textsc{Train}$(N)$ fails}
    \STATE $N_{\text{fail}} \gets N$
    \IF{$N/2 < F$} \STATE \textbf{return} $0$ \ENDIF
    \STATE $N \gets \mathrm{round}_S(N/2)$ if $N/2 \geq S$, else $\mathrm{round}_F(N/2)$ \COMMENT{halving: coarse grid above $S$, fine grid below}
  \ENDWHILE
  \STATE $N_{\text{ok}} \gets N$
\ENDIF
\STATE $lo \gets$ the smallest multiple of $F$ above $N_{\text{ok}}$; $hi \gets$ the largest multiple of $F$ below $N_{\text{fail}}$ \COMMENT{binary search on the $F$ grid}
\WHILE{$lo \leq hi$}
  \STATE $m \gets$ the multiple of $F$ nearest to $(lo + hi)/2$
  \IF{\textsc{Train}$(m)$ passes} \STATE $N_{\text{ok}} \gets m$; $lo \gets m + F$ \ELSE \STATE $hi \gets m - F$ \ENDIF
\ENDWHILE
\STATE \textbf{return} $N_{\text{ok}}$
\end{algorithmic}
\end{algorithm}
 
\end{document}